\documentclass{article}

\usepackage{microtype}
\usepackage{graphicx}
\usepackage{enumitem}
\usepackage{subcaption}
\usepackage{booktabs} 

\usepackage{hyperref}

\usepackage[preprint]{icml2026}

\usepackage{amsmath}
\usepackage{amssymb}
\usepackage{mathtools}
\usepackage{booktabs}
\usepackage{threeparttable}
\usepackage{graphicx}
\usepackage{amsthm}
\usepackage{multirow}
\usepackage[table]{xcolor}
\usepackage{xspace}
\usepackage{makecell}   
\usepackage[frozencache,cachedir=minted-cache]{minted}
\usepackage{listings}
\usepackage[most]{tcolorbox}
\tcbuselibrary{listingsutf8}
\usepackage{booktabs} 
\usepackage{threeparttable}

\definecolor{rowgray}{gray}{0.85}
\definecolor{rowlightblue}{RGB}{215, 230, 255} 

\usepackage[capitalize,noabbrev]{cleveref}

\theoremstyle{plain}

\theoremstyle{definition}

\theoremstyle{remark}

\definecolor{gemblue}{RGB}{225, 240, 255}
\usepackage[textsize=tiny]{todonotes}

\icmltitlerunning{\textsc{AOrchestra}: Automating Sub-Agent Creation for Agentic Orchestration}

\newcommand{\our}{\textsc{AOrchestra}\xspace}

\begin{document}

\twocolumn[
  \icmltitle{\textsc{AOrchestra}: Automating Sub-Agent Creation for Agentic Orchestration}

  \begin{icmlauthorlist}
      \icmlauthor{Jianhao Ruan\textsuperscript{*}}{1,3}
      \icmlauthor{Zhihao Xu\textsuperscript{*}}{2}
      \icmlauthor{Yiran Peng}{1}
      \icmlauthor{Fashen Ren}{3}
      \icmlauthor{Zhaoyang Yu}{1} \\
      \icmlauthor{Xinbing Liang}{4}
      \icmlauthor{Jinyu Xiang}{3}
      \icmlauthor{Yongru Chen}{3}
      \icmlauthor{Bang Liu}{5}
      \icmlauthor{Chenglin Wu}{1}
      \icmlauthor{Yuyu Luo}{3}
      \icmlauthor{Jiayi Zhang}{1,3}
    \end{icmlauthorlist}

\icmlaffiliation{1}{DeepWisdom}
\icmlaffiliation{2}{RUC}
\icmlaffiliation{3}{HKUST(GZ)}
\icmlaffiliation{4}{ECNU}
\icmlaffiliation{5}{UdeM \& Mila}

  \icmlcorrespondingauthor{Yuyu Luo}{yuyuluo@hkust-gz.edu.cn}
  \icmlcorrespondingauthor{Jiayi Zhang}{jzhang361@connect.hkust-gz.edu.cn}
  \vskip 0.3in
]



\printAffiliationsAndNotice{}  


\begin{abstract}


Language agents have shown strong promise for task automation. 
Realizing this promise for increasingly complex, long-horizon tasks has driven the rise of a sub-agent-as-tools paradigm for multi-turn task solving. 
However, existing designs still lack a \emph{dynamic abstraction} view of sub-agents, thereby hurting adaptability.
We address this challenge with a unified, framework-agnostic agent abstraction that models any agent as a tuple $\langle Instruction, Context,Tools,Model \rangle$. This tuple acts as a compositional recipe for capabilities, enabling the system to spawn specialized executors for each task on demand. 
Building on this abstraction, we introduce an agentic system \textsc{AOrchestra}, where the central orchestrator concretizes the tuple at each step: it curates task-relevant context, selects tools and models, and delegates execution via on-the-fly automatic agent creation.
Such designs enable reducing human engineering efforts, and remain framework-agnostic with plug-and-play support for diverse agents as task executors. It also enables a controllable performance–cost trade-off, allowing the system to approach Pareto-efficient.
Across three challenging benchmarks (GAIA, SWE-Bench, Terminal-Bench), \textsc{AOrchestra} achieves 16.28\% relative improvement against the strongest baseline when paired with Gemini-3-Flash. The code is available at: \url{https://github.com/FoundationAgents/AOrchestra  }

\end{abstract}

\section{Introduction}
Humans handle complex, long-horizon work via collective intelligence and the ability to coordinate~\cite{gao2025evolvesurvey,DBLP:journals/corr/abs-2510-23587,DBLP:journals/corr/abs-2510-17586}. As today's agents are pushed toward similarly complex and multi-turn tasks~\cite{yao2024tau, zhang2025autoenv, xu2026unlocking}, a well-designed agentic system becomes a vital way to scale performance beyond a single model~\cite{liu2025advances}.

\begin{figure}[t!]
    \centering
    \includegraphics[width=\linewidth]{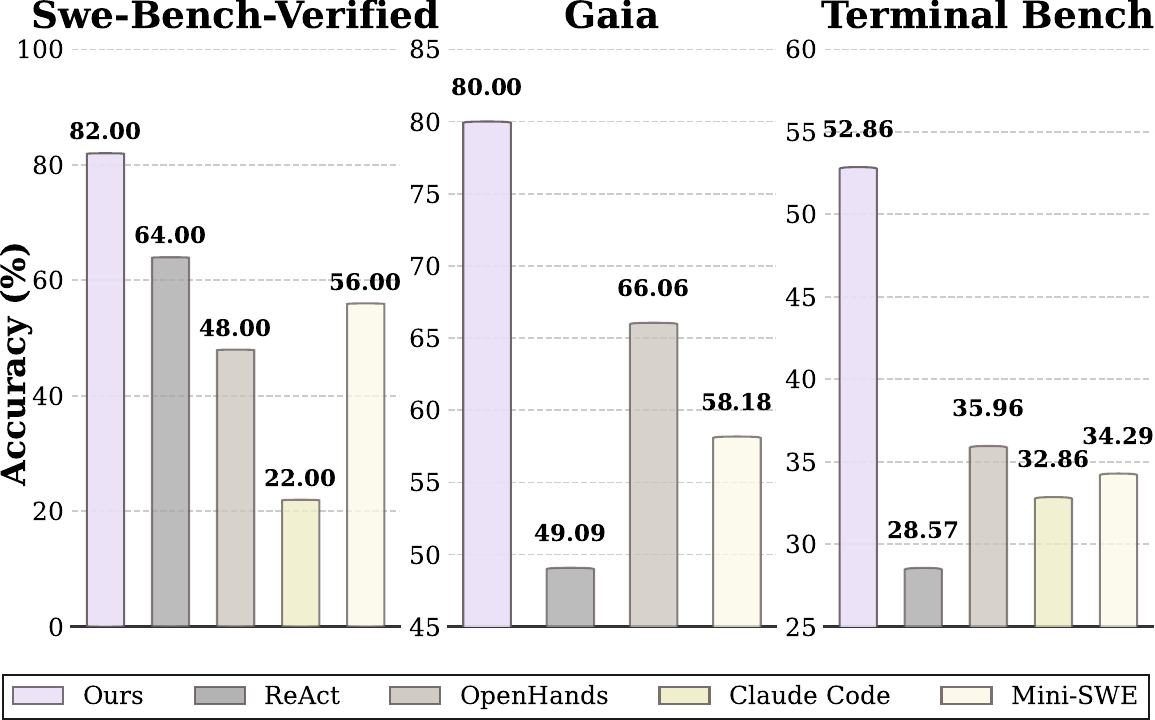}
    \caption{Overall performance on three challenging agentic benchmarks (GAIA, Terminal-Bench-2, SWE-Bench-Verified) paired with Gemini-3-Flash when comparing \our against other popular agentic frameworks.}
    \label{fig:performance}
\end{figure}

To cope with increasingly complex scenarios, early attempts rely on fixed coordination workflows or multi-agent systems~\cite{hong2023metagpt, hu2025owl, DBLP:journals/corr/abs-2510-17586}. 
While multi-agent collaboration can improve task decomposition, in open-ended environments it often incurs substantial coordination overhead and provides limited control over context routing, leading to either noisy over-sharing or harmful omission of critical information, which makes robust long-horizon execution difficult~\cite{gao2025single}.

\begin{figure*}[htbp]
    \centering
    \includegraphics[width=\linewidth]{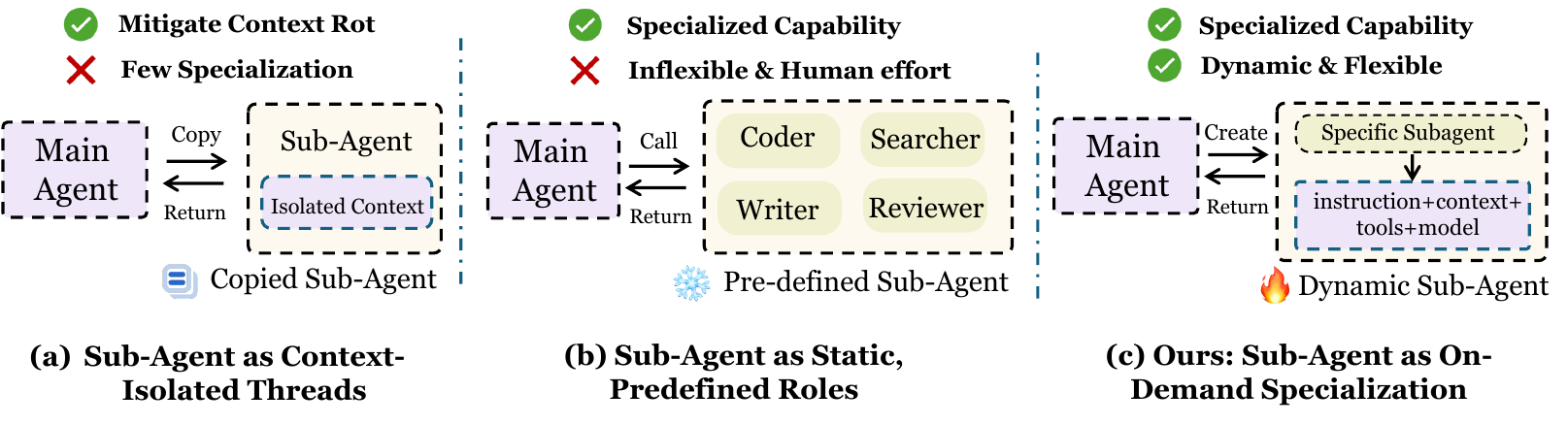}
    \caption{Comparison of sub-agent-as-tools approaches. \textbf{(a) Sub-agents as context-isolated threads} mitigate context rot but lack on-demand specialization. \textbf{(b) Sub-agents as static roles} provide specialized capabilities but are inflexible, leave coverage gaps, and require heavy human engineering. \textbf{(c) Our Sub-agents as on-demand specialization} concretizes a unified 4-tuple abstraction \textsc{(Instruction, Context, Tools, Model)} to enable creating tailored executors on the fly.}
    \label{fig:diff}
\end{figure*}

More recent approaches therefore move toward a more practical sub-agent-as-tools paradigm, where a main agent (orchestrator) delegates a task to a sub-agent via an explicit tool call. 
Yet existing designs still lack flexibility in practice and often degenerate into two limited patterns, which are shown in Figure~\ref{fig:diff}:
(1) \textit{Sub-agents as context-isolation threads.} Systems such as ~\citet{schroeder2025thread, sun2025scaling} primarily treat sub-agents as isolated context threads, aiming to prevent context rot~\citep{hong2025context}. However, in real-world tasks, subtasks often require specialized capabilities. Therefore, such systems fail to fully realize the potential of specialized sub-agents. (2) \textit{Sub-agents as static roles.} 
Systems such as ~\citet{anthropic_claude_code_subagents, li2025chain} treat each sub-agent as a static role, and their capabilities or their coordination patterns are typically hard‑wired. A pre-defined set of sub‑agents cannot cover the dynamically emerging variety of subtasks in open environments. Besides, it relies on heavy human engineering, making the system difficult to adapt to various environments.

In this paper, we introduce \our{}, an agentic framework designed to tackle long-horizon and complex tasks. Our core insight lies in treating sub-agents through the lens of \textbf{on-demand specialization}, as illustrated in Figure~\ref{fig:diff}(c). 
We posit that a sub-agent should be viewed as a flexible abstraction unit rather than a predefined, fixed role. This approach enables the system to instantiate tailored sub-agents at runtime by dynamically composing their capabilities to meet specific task demands—an essential feature.
Concretely, \emph{any} agent can be described as an instantiable unit via a unified four-tuple: \textsc{(instruction, context, tools, model)}. This specialization is organized around two complementary axes essential for an agent's task solving: (1) \emph{Working memory (instruction, context):} what the agent must achieve and what evidence it should condition on. Notably, the context attribute is designed to inject only the most relevant information for the current sub-task, filtering out potentially distracting details. (2) \emph{Capabilities (tools, model):} what the agent is empowered to do to accomplish that objective. By composing specific tools and models on a per-subtask basis, we endow each sub-agent with precise, task-specific functionality. Together, this 4-tuple design enables an automatic specialized sub-agent for each task.

Building on this on-demand specialization view, we further introduce a dedicated orchestrator that operates directly over the four-tuple interface to automatically create tailored sub-agents on the fly. It does not execute any tasks and focuses exclusively on orchestration, where we define it as dynamically decomposing the overall objective into the next subtask, creating and delegating a specialized tailored sub-agent for task execution via explicit tool calls.
This decoupling design offers several key advantages. 
First, this dynamic creation allows each sub-agent to be customized with unique capabilities and a clean working context, significantly improving task execution accuracy. 
Second, the orchestrator remains agnostic to the internal implementation of sub-agents, making them fully pluggable. 
Third, the orchestrator can be trained or learned from interactive experience.  
This ranges from basic skills for agent creation to advanced features like adaptive model selection, achieving an optimal balance between cost and performance.

Through extensive experiments, we demonstrate \our{} achieves stronger performance and broader generalization in open‑world settings.
We first evaluate our framework in a training-free setting on three challenging agentic benchmarks: Terminal-Bench 2.0~\cite{tbench_2025} (bash environment), SWE-Bench~\cite{jimenez2023swe} (coding environment), and GAIA~\cite{mialon2023gaia} (digital world environment). As shown in Figure~\ref{fig:performance}, our method consistently outperforms both representative sub-agent orchestration approaches ~\cite{anthropic_claude_code_subagents} and widely used agent frameworks~\cite{wang2024openhands, yang2024sweagent} across all benchmarks. In particular, our framework achieves a 16.28\% improvement when paired with \texttt{Gemini3-Flash}, validating the superiority of our orchestration model in complex, long-horizon tasks. 
Importantly, \our naturally supports learning the orchestration policy from experience.
We instantiate this in two ways: (1) we apply supervised fine-tuning to improve the Orchestrator's subtask decomposition and 4-tuple synthesis, leading to better orchestration quality by +11.51\% pass@1 on GAIA and
(2) we leverage in-context learning to optimize cost-aware routing, which improves GAIA pass@1 by +3.03\% while reducing average cost by 18.5\%, resulting in a more favorable cost--performance Pareto frontier.

Overall, our contributions are:
\begin{itemize}
    \item We propose \our, an orchestrator-centric agentic system that treats sub-agents as \emph{dynamically creatable} executors via a unified 4-tuple interface \textsc{(Instruction, Context, Tools, Model)}, enabling on-demand specialization with task-sufficient context and explicit capability control.
    \item \our achieves strong training-free performance on Terminal-Bench 2.0, SWE-Bench-Verified, and GAIA, consistently outperforming popular agentic systems. We achieve 16.28\% relative improvement against the strongest baseline when paired with Gemini-3-Flash.  
    \item We show the orchestration policy is learnable under this design from two complementary angles: supervised fine-tuning improves basic task orchestration (+11.51\%), and cost-aware routing via in-context learning yields favorable cost--performance Pareto trade-offs (reducing average cost by 18.5\%).
\end{itemize}

\section{Related Work}
\paragraph{Multi-Agent Systems}
Inspired by collaborative problem solving, early efforts propose multi-agent systems (MAS) to enhance the task-solving capability of language models~\cite{zhang2025agentorchestra, DBLP:journals/corr/abs-2505-07437,wu2024autogen, shi2025aime, gao2025single, zhu2025oagents, fang2025cognitive, zhang2024aflow,DBLP:conf/icml/Li0FXC0L25}.
For example, MetaGPT~\cite{hong2023metagpt} organizes agents into a structured software-development workflow, where specialized roles (e.g., product manager, architect) collaborate via predefined communication protocols.
OWL~\cite{hu2025owl} adopts a planner-worker workflow to improve transfer and generalization by modularizing domain-agnostic planning and domain-specific execution.
Despite their effectiveness, most MAS typically rely on a fixed workflow, leading to rigidity.
Although AutoAgents~\cite{chen2023autoagents} proposes building different multi-agent systems for each task, they still rely on a fixed workflow to accomplish this.
This motivates a growing shift toward the \emph{sub-agents-as-tools} paradigm, and we will list related works in the next part~\cite{gao2025evolvesurvey, gao2025single}. 
\our{} follows the latter and further emphasizes orchestration-centric, dynamic sub-agent creation without relying on a specific human-designed workflow. 

\paragraph{Sub-Agent as Tools}
This approach involves a primary agentic model invoking a sub-agent in a tool-like manner to solve problems~\cite{li2025chain, su2025toolorchestra, grand2025self,DBLP:journals/tkde/LiuSLMJZFLTL25}. For example, THREAD~\cite{schroeder2025thread} enables the recursive spawning of sub-agents to address decomposed subproblems. Similarly, Context-Folding~\cite{sun2025scaling} proposes branching for a subtask and then folding it back by compressing intermediate steps into a concise summary, thereby managing context.
However, these methods do not treat sub-agents as fully specialized agents, leading to their insufficient utilization. 
Other practical systems, such as Claude Code~\cite{anthropic_claude_code_subagents}, support sub-agents that operate within isolated context windows with custom system prompts and tool permissions. Yet, these sub-agents are typically configured as fixed specialists and still require manual design.
\our{} addresses these limitations by treating each sub-agent as a dynamic unit and proposes an orchestration-centric agentic system that proactively and dynamically creates such sub-agents on demand.

\section{Methodology}

\begin{figure*}[t]
    \centering
    \includegraphics[width=\textwidth]{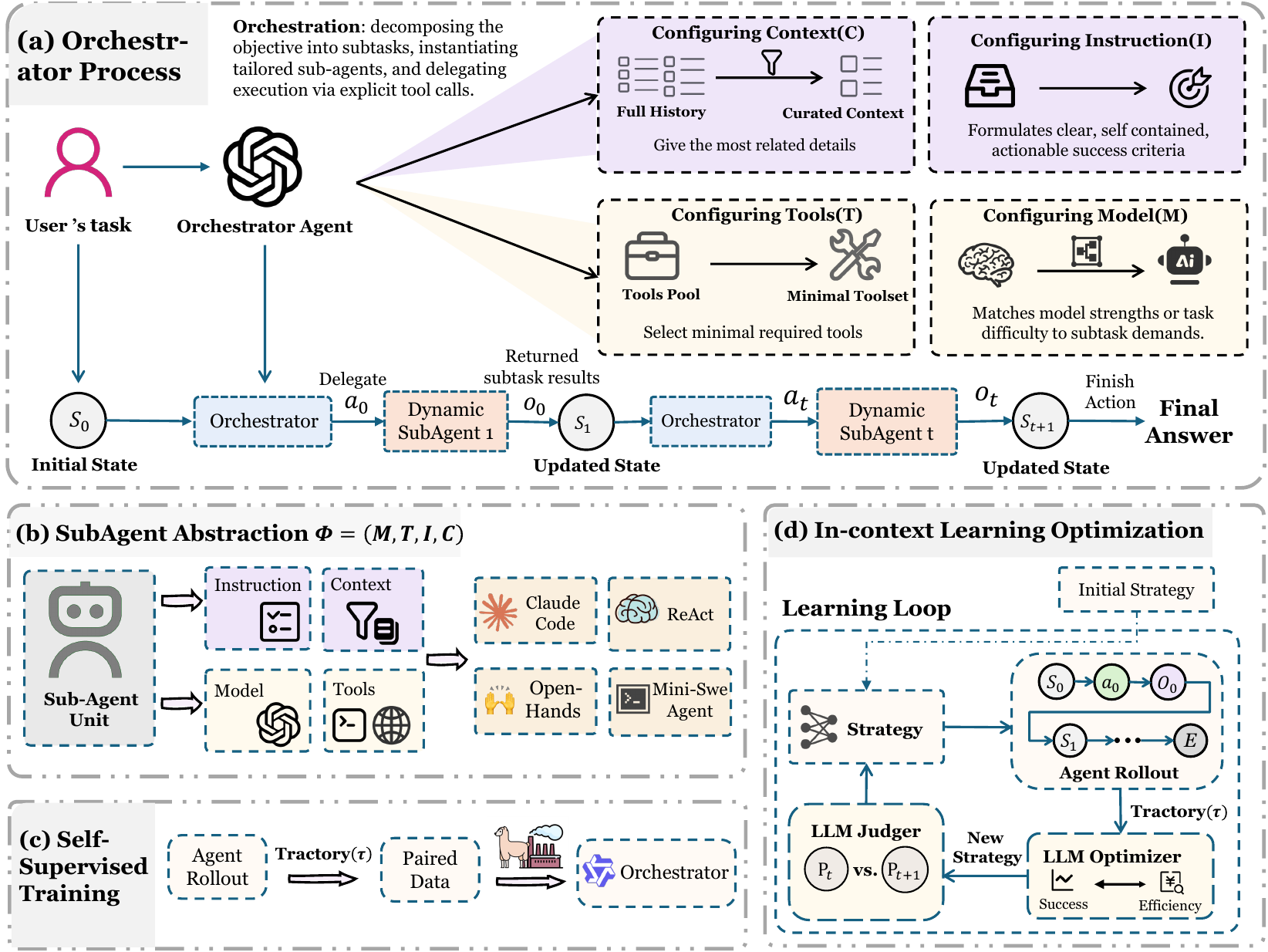}
    \caption{Overall design of our proposed agentic framework, \our{}, for complex, long-horizon tasks. The orchestrator solves a user task by repeatedly delegating subtasks to on-the-fly instantiated sub-agents, each defined by a unified four-tuple $(I,C,T,M)$. The orchestrator is learnable and can improve its decomposition, context routing, and capability allocation from past experience.}
    \label{fig:piepine}
\end{figure*}

In this section, we first formalize the problem in Section~\ref{sec:formulation}. Next, we elaborate on the design of \our in Section~\ref{sec:method_design}. Finally, we introduce the process for training a dedicated orchestrator in Section~\ref{sec:learnable}. Figure~\ref{fig:piepine} provides an overview of our methodology.

\subsection{Problem Formulation}
\label{sec:formulation}
In this paper, we mainly focus on solving complex agentic tasks. The agentic system solves a user goal $G$ through multi-step interaction with an environment.
The environment exposes an \emph{environment-level} action space $\mathcal{A}_{\text{env}}$ (e.g., shell commands, web operations, code edits) and returns feedback such as observations, tool outputs, and error messages.
An interaction trajectory for a task can be therefore defined as:
\[
\tau = (s_0, a_0, o_0, s_1, a_1, o_1, \dots, s_T),
\]
where $s_t \in \mathcal{S}$ denotes the system state at step $t$ (including accumulated history, intermediate results, and environment feedback), $a_t$ is the action taken at step $t$, and $o_t \in \mathcal{O}$ is the returned observation.
The system evolves according to a state-transition function
\[
s_{t+1} = \delta(s_t, a_t, o_t),
\]
where $\delta: \mathcal{S} \times \mathcal{A}_{\text{env}} \times \mathcal{O} \rightarrow \mathcal{S}$ maps the current state, action, and observation to the next state by incorporating newly returned information into the system's internal state.

\paragraph{Sub-agent-as-tools view.}
We focus on the sub-agent-as-tools paradigm, where a \emph{main agent} (orchestrator) can either act in the environment directly or delegate a subtask to a sub-agent as a tool call.
Accordingly, the orchestrator operates over a \emph{system-level} action space that typically includes three types of actions:
(1) environment actions $u \in \mathcal{A}_{\text{env}}$,
(2) delegation actions (\texttt{Delegate}$(\cdot)$) that invoke a sub-agent to execute,
and (iii) termination (\texttt{Finish}).
We denote this generic orchestration action space as
\[
\mathcal{A}_{\text{orch}} \supseteq \mathcal{A}_{\text{env}} \ \cup\ \{\texttt{Delegate}(\cdot),\ \texttt{Finish}(y)\}.
\]
Different systems mainly differ in how \texttt{Delegate} is parameterized (e.g., delegating with only context vs.\ delegating to a fixed set of roles), and in whether the orchestrator itself also performs environment actions.

Our objective is to maximize task success, optionally trading off execution cost:
\[
\max_{\pi}\ \mathbb{E}\Big[\mathbf{1}\{\text{Success}(G)\}-\lambda\cdot \text{Cost}(\tau)\Big],
\]
where $\pi$ is the orchestrator policy, $\text{Cost}(\tau)$ may include token usage, tool calls, latency, or monetary cost, and $\lambda$ controls the cost--performance trade-off.

\subsection{\our{} }
\label{sec:method_design}

\paragraph{A unified four-tuple agent abstraction.}
\our{} models \emph{both} the main agent and sub-agents under a unified framework-agnostic abstraction.
We define an agent instance as an instantiable four-tuple
\[
\Phi = (I, C, T, M),
\]
where $I$ is the task instruction specifying the current objective and success criteria,
$C$ is the curated working context the agent conditions on,
$T$ is the tool set defining the agent's action space,
and $M$ is the underlying model that interacts with the environment.
This abstraction explicitly separates two complementary axes that require \textit{specialization}:
\emph{working memory} $(I,C)$ and \emph{capabilities} $(T,M)$.
Notably, we do not view a sub-agent as a static entity, but rather as a \emph{dynamic} unit that can be parametrized and created at runtime. 

The main agent (orchestrator) can also be represented by a tuple
$\Phi^{\text{main}}=(I^{\text{main}},C^{\text{main}},T^{\text{main}},M^{\text{main}})$.
The difference is that $T^{\text{main}}$ exposes \emph{system tools} for orchestration (e.g., \texttt{Delegate}, \texttt{Finish}) rather than environment tools in $\mathcal{A}_{\text{env}}$.

\paragraph{Action Space of Orchestrator.}
Building on this abstraction, \our{} decouples orchestration from execution.
The orchestrator in \our{} never directly takes environment actions in $\mathcal{A}_{\text{env}}$.
Instead, it operates only the two following actions:
\[
\mathcal{A}_{\our} = \{\texttt{Delegate}(\Phi),\ \texttt{Finish}(y)\}.
\]
At step $t$, the orchestrator samples an action $a_t \in \mathcal{A}_{\our}$.
If $a_t=\texttt{Delegate}(\Phi_t)$, it spawns an executor $A(\Phi_t)$ to execute the subtask and returns an observation $o_t$.
If $a_t=\texttt{Finish}(y)$, the interaction terminates with the final answer $y$.
Returned observations are integrated into the next state via $s_{t+1}=\delta(s_t,a_t,o_t)$.

\paragraph{Implementation of \texttt{Delegate} and \texttt{Finish}.}
We implement \our{} with two system tools available to the Orchestrator: \texttt{Delegate} and \texttt{Finish}.
\texttt{Delegate} takes $\Phi_t=(I_t,C_t,T_t,M_t)$ as arguments and instantiates an executor accordingly.
The executor runs with model $M_t$, is restricted to the tool set $T_t$, and conditions only on $(I_t,C_t)$.
It returns a structured observation $o_t$ to the Orchestrator, typically including (i) a concise result summary, (ii) relevant artifacts (e.g., files, references), and (iii) error messages or logs if execution fails.
\texttt{Finish} terminates the interaction and outputs the final response $y$.

\paragraph{Advantages of \our{}}
Our proposed \our{} offers several key advantages.
First, it dynamically equips each sub‑agent with tailored capabilities on demand, which substantially improves the accuracy of task execution. Unlike prior works~\cite{sun2025scaling, anthropic_claude_code_subagents}, the orchestrator deliberately provides well‑structured context for the sub‑agent to use. As shown later in Section~\ref{sec:adv1}, this careful context management enhances the model’s ability to solve tasks.
Second, the orchestrator operates solely on a four‑tuple abstraction and remains independent of the internal implementation of sub‑agents. This flexibility allows us to employ various designs for sub‑agents, such as a simple React approach~\cite{yao2022react} or a mini‑SWE agent.
Third, the orchestrator can learn from extensive experience. We will then detail this in Section~\ref{sec:learnable}. These learnable aspects include basic task orchestration skills (i.e., what to do, what to condition on, and which tool to use) as well as advanced features (e.g., adaptive model routing, where the goal might be to balance performance and cost by selecting the most suitable model).


\subsection{Learnable Orchestrator}
\label{sec:learnable}

With $\mathcal{A}_{\our}={\texttt{Delegate}(\Phi),\texttt{Finish}(y)}$, the orchestration task can be expressed as learning a policy over structured actions:
\[
\pi_\theta(a_t \mid s_t),\quad a_t\in\mathcal{A}_{\our}.
\]

In this paper, learning mainly focuses on the two following complementary dimensions:
Since the delegation parameters $\Phi_t=(I_t,C_t,T_t,M_t)$ are explicitly available, learning can focus on two complementary dimensions:
(i) \textbf{Task orchestration}, which determines what to do, what context to use, and which tools to employ.
(ii) \textbf{Model routing}, which selects $M_t$ (the model to call) to balance performance and cost.
In the following, we detail these two learning paradigms.

\paragraph{Supervised fine-tuning (SFT) for task orchestration.}
Given expert orchestration trajectories $\{(s_t,a_t^\star)\}$, we fine-tune the Orchestrator by behavior cloning:
\[
\theta^\star = \arg\max_{\theta}\ \sum_{t} \log p_{\theta}(a_t^\star \mid s_t),
\]
where $a_t^\star$ is the expert action (either $\texttt{Delegate}(\Phi_t^\star)$ or $\texttt{Finish}(y^\star)$).
In our setup, SFT primarily distills \emph{task orchestration}: improving subtask decomposition and the synthesis of $(I_t,C_t,T_t)$, i.e., producing better working memory, and more appropriate tool subsets for each step.
We would like to note that in this work, we prioritize showing the potential of training a specialized orchestrator, thus employing a straightforward SFT approach. Note that others can employ any training methods like GRPO~\cite{shao2024deepseekmath} to improve the task orchestration capability.

\paragraph{Iterative In-context Learning for Cost-aware Orchestration.}
Beyond parameter updates, we also optimize orchestration \emph{without} changing model weights by learning the Orchestrator's \emph{instruction} (prompt) through iterative interaction.
Concretely, we treat the Orchestrator instruction $I^{\text{main}}$ as the learnable object and run \our{} in the environment to collect trajectories
$\tau=\{(s_t,a_t,o_t)\}_{t=0}^{T}$ together with outcome metrics, including task performance and execution cost.
An optimization model then analyzes these trajectories and proposes prompt edits $\Delta I$ to update the instruction:
\[
I^{\text{main}}_{k+1} = \textsc{Optimize}\bigl(I^{\text{main}}_{k},\ \tau_k,\ \text{Perf}(\tau_k),\ \text{Cost}(\tau_k)\bigr),
\]
where $k$ indexes optimization rounds.
By repeatedly rolling out the updated Orchestrator in the environment for $N$ rounds, this process improves cost-aware orchestration behavior (e.g., model compiler/routing decisions and tool usage patterns) and aims to discover Pareto-efficient trade-offs between performance and cost.

\section{Experiments}

\subsection{Experiment Setup}

\begin{table*}[!t]
\centering
\small
\setlength{\tabcolsep}{5pt}
\renewcommand{\arraystretch}{1.1}
\begin{threeparttable}
\caption{Comparison between \our and baseline agentic systems on GAIA, Terminal-Bench 2.0, and SWE-Bench-Verified under various models. The best results are in \textbf{bold}.}
\label{tab:main_results}
\begin{tabular}{l l cc cc cc c}
\toprule
\textbf{Methods} & \textbf{Model Setup} 
& \multicolumn{2}{c}{\textbf{GAIA}} 
& \multicolumn{2}{c}{\textbf{Terminal-Bench 2.0}} 
& \multicolumn{2}{c}{\textbf{SWE-Bench-Verified}}
\\
\cmidrule(lr){3-4}\cmidrule(lr){5-6}\cmidrule(lr){7-8}
& 
& \textbf{Pass@1} & \textbf{Pass@3}
& \textbf{Pass@1} & \textbf{Pass@3}
& \textbf{Pass@1} & \textbf{Pass@3}
& \textbf{Avg. Pass@1}\\
\midrule

\multirow[c]{3}{*}{ReAct}
& Gemini-3-Flash     & 49.09 & 66.06 & 28.57 & 47.14 & 64.00 & 82.00 & 47.22 \\
& DeepSeek-V3.2      & 46.70 & 71.51 & 20.00 & 32.86 & 48.00 & 87.00 & 38.23 \\
& Claude-4.5-haiku   & 47.88 & 62.42 & 20.00 & 37.14 & 63.00 & 87.00 & 43.62 \\
\cmidrule(lr){1-9}

\multirow[c]{3}{*}{OpenHands}
& Gemini-3-Flash     & 66.06 & 72.73 & 31.43 & 51.43 & 48.00 & 66.00 & 48.49 \\
& DeepSeek-V3.2      & 63.64 & 72.12 & 21.43 & 35.71 & 60.00 & 75.00 & 48.35 \\
& Claude-4.5-haiku   & 54.55 & 61.21 & 12.85 & 25.71 & 68.00 & 83.00 & 45.13 \\
\cmidrule(lr){1-9}

\multirow[c]{3}{*}{Mini-SWE}
& Gemini-3-Flash     & 58.18 & 68.48 & 34.29 & 50.00 & 56.00 & 85.00 & 49.49 \\
& DeepSeek-V3.2      & 50.30 & 63.63 & 30.00 & 48.57 & \textbf{84.00} & \textbf{89.00} & 54.76 \\
& Claude-4.5-haiku   & 40.61 & 60.00 & 24.29 & 28.57 & 44.00 & 83.00 & 36.30 \\
\cmidrule(lr){1-9}

\multirow[c]{2}{*}{Claude Code}
& Gemini-3-Flash     & -- & -- & 32.86 & 48.57 & 22.00 & 42.00 & 27.43 \\
& Claude-4.5-haiku   & -- & -- & 34.29  & 45.71 & 25.00 & 41.00 & 29.65 \\
\midrule

\multirow[c]{3}{*}{\our}
& Gemini-3-Flash     & \textbf{80.00} & \textbf{86.06} & \textbf{52.86} & \textbf{57.14} & 82.00 & 86.00 & \textbf{71.62} \\
& DeepSeek-V3.2      & 67.87 & 80.00 & 31.43 & 42.86 & 76.00 & 82.00 & 58.43 \\
& Claude-4.5-haiku   & 60.61 & 73.90 & 35.71 & 45.71 & 70.00 & 84.00 & 55.44 \\
\bottomrule
\end{tabular}
\end{threeparttable}
\end{table*}

\textbf{Benchmarks.} We evaluate our method on three challenging agentic benchmarks that span diverse interactive settings: (1) \textbf{Terminal-Bench 2.0}~\cite{tbench_2025}, which places agents in a Linux terminal with an interactive Bash shell, requiring them to execute command-line operations to complete multi-step real-world tasks; (2) \textbf{SWE-Bench-Verified}~\cite{jimenez2023swe}, which assesses software engineering on real GitHub projects, where agents must localize bugs, implement patches, and satisfy the provided test suites under realistic coding environment; and (3) \textbf{GAIA}~\cite{mialon2023gaia} validation set, a generalist benchmark that tests an agent's ability to solve real-world tasks requiring multi-step reasoning and tool use. We report pass@1 and pass@3 for all benchmarks respectively. 
We report more details about how we use these datasets for evaluation in Appendix~\ref{app:dataset}.
We also detail the tools we used for each benchmark in Appendix~\ref{app:action_space}.

\textbf{Model \& Baselines.}
We compare our method against representative frameworks: 
(1) \textbf{ReAct}~\cite{yao2022react}, a simple single-agent system directly build on ReAct that interleaves reasoning and actions;
(2) \textbf{OpenHands}~\cite{wang2024openhands}, a commonly-used open agent platform for solving diverse real-world tasks;
(3) \textbf{mini-SWE-agent}~\cite{yang2024sweagent}, a minimalistic coding agent designed to solve GitHub issues and more;
and (4) \textbf{Claude Code}~\cite{anthropic_claude_code_subagents}, a production-grade agentic CLI that supports spawning pre-defined sub-agents for task decomposition and context isolation.
For each agentic system, we employ the following frontier language models, including two strong models (\texttt{Gemini-3-Flash} and \texttt{DeepSeek-V3.2}~\cite{liu2025deepseek}) and a smaller model (\texttt{Claude-4.5-haiku}). We report the implementations of baselines in Appendix~\ref{app:baseline_imp}.

\textbf{Implementation.} 
Across all experiments, we set $max\_attempt=10$ for the orchestrator and $max\_step=50$ for the sub-agent. We set $max\_step=500$ for all baselines for a fair comparison.
For the training-free setting, we detail our designs in Appendix~\ref{app:prompt_used}, which includes all prompts for the orchestrator and sub-agent we use across three benchmarks. After the sub-agents complete their tasks, a reviewer LLM reviews the execution trace and summarizes the core insights.


For SFT training, we fine-tune Qwen3-8B~\cite{yang2025qwen3} to improve its orchestration capability in non-thinking mode. We use TaskCraft~\cite{shi2025taskcraft} as the seed dataset and employ Gemini-3-Flash to collect 2K orchestration trajectories for SFT training. During SFT, we perform full-parameter fine-tuning under LLamaFactory framework~\cite{zheng2024llamafactory} for 2 epochs with a learning rate of 1e-5, with more details in Appendix~\ref{app:sft_hyper}.

For in-context learning, we use \texttt{Claude Sonnet 4.5} as an optimization model to iteratively update the Orchestrator instruction.
We run 5 optimization rounds; in each round, we collect 6 interaction trajectories for analysis.
After each round, the prompt that achieves the best cost--performance trade-off (highest performance with lower cost) is selected for initializing the next round.

\subsection{Main Results}

\label{sec:main_results}
Table~\ref{tab:main_results} presents the main results of \our compared to baseline agentic systems on three benchmarks (GAIA, Terminal-Bench 2.0, and SWE-Bench-Verified), evaluated by pass@1/pass@3 metric. For \our, we use \texttt{Gemini-3-Flash} as the orchestrator and use only one model as sub-agent choices here for comparison.
Overall, we find that \our consistently outperforms the baselines on all environments. \our{} outperforms the best baselines by an average of 22.13\% pass@1 with \texttt{Gemini-3-Flash} across three benchmarks.

\paragraph{GAIA Results}
GAIA measures the ability of a general-purpose agent to solve real-world tasks, such as multi-hop searching, file processing, and multimodal operations. In such an environment, \our achieves the strongest performance against all baselines.
Specifically, with \texttt{Gemini-3-Flash} as both the orchestrator and the sub‑agent model, \our achieves 80.00 pass@1 and 86.06 pass@3, which represents the best performance among all baselines. Under the same \texttt{Gemini-3-Flash} backbone, \our raises pass@1 by 13.94 points absolute over the strongest baseline framework, \texttt{OpenHands}, increasing the result from 66.06 to 80.00. Even with a less powerful model \texttt{Claude‑4.5‑haiku} as the sub-agent model, it still attains 60.61 pass@1 on GAIA, confirming that the observed improvements are not confined to the most capable model configuration. We do not evaluate Claude Code for GAIA because it is designed as a production-level coding agent, and thus its corresponding result is left blank. We present a case study on GAIA in Appendix~\ref{app:case_study}.

\paragraph{Terminal-Bench 2.0 Results}
Terminal-Bench assesses an agent’s ability to operate in computer terminal environments inspired by real-world workflows. 
On this benchmark, \our with \texttt{Gemini-3-Flash} achieves 52.86 pass@1 and 57.14 pass@3. This is an absolute improvement of 64.29 points in pass@1 over the strongest baseline in Table~\ref{tab:main_results}, \texttt{Mini-SWE} with 34.29 pass@1.
Beyond the \texttt{Gemini-3-Flash} setting, \our remains competitive under other backbones, with performance that is comparable to or better than specialized coding agentic systems such as \texttt{Claude Code}.

\paragraph{SWE-Bench-Verified Results}
SWE-Bench-Verified evaluates an agent’s ability to resolve real issues in open-source repositories by producing code patches that pass the provided tests.
On this benchmark, \our achieves strong performance across backbones and is competitive with the best baseline systems. With \texttt{Gemini-3-Flash}, \our reaches 82.00 pass@1 and 86.00 pass@3, outperforming \texttt{ReAct} and \texttt{OpenHands} under the same model setting. Compared with \texttt{Mini-SWE}, which is designed for software tasks, \our remains competitive, and it consistently achieves over 70.00 pass@1 across all three model backbones.

\subsection{Advantage Analysis of \our}
\label{subsec:adv}

In this section, we present analyses that demonstrate the benefits of \our for dynamically creating specialized sub‑agents, particularly in terms of working memory and capabilities. Overall, our findings indicate that explicitly passing context from the orchestrator to sub‑agents yields performance gains (Sec.~\ref{sec:adv1}). We also show that selecting different models for different tasks can achieve a cost‑performance Pareto (Sec.~\ref{sec:adv2}), and that diverse implementations of sub‑agents consistently contribute to overall improvement (Sec.~\ref{sec:adv3})

\subsubsection{Advantage 1: Context Sharing}
\label{sec:adv1}

\begin{table}[t]
\centering
\small
\setlength{\tabcolsep}{6pt}
\renewcommand{\arraystretch}{1.15}
\caption{Context-control ablation for sub-agent invocation. We isolate the effect of context inheritance by only changing the \texttt{Context} field passed to sub-agents, while keeping the sub-agent model, tools, and system prompt identical across settings.}
\label{tab:context_control}
\begin{tabular}{l c c c c}
\toprule
\textbf{Setting} & \textbf{Level 1} & \textbf{Level 2} & \textbf{Level 3} & \textbf{Avg. } \\
\midrule
No-Context    & 89.47 & 81.48 & 75.00 & 86.00 \\
Full-Context  & 94.74 & 77.78 & 75.00 & 84.00 \\
\textbf{Ours} & \textbf{100.00} & \textbf{88.89} & \textbf{75.00} & \textbf{96.00} \\
\bottomrule
\end{tabular}
\end{table}

In \our{}, the orchestrator explicitly and dynamically passes curated context to each created sub‑agent. To evaluate the effectiveness of this design, we compare it with two variants: \textbf{No-Context}, where each sub-agent only receives a task instruction, and \textbf{Full-Context}, where each sub-agent inherits all context from the main agent. Note that these two approaches are also commonly used in prior systems~\cite{sun2025scaling, anthropic_claude_code_subagents}. 
Here, we conduct analysis on GAIA, and sample 50 samples from the validation set.

Table~\ref{tab:context_control} indicates that it is necessary to regard context as an important component of sub-agent and abstract it into one of the four tuples.
In particular, we find that No‑Context fails due to the lack of critical execution traces and fine‑grained cues from previous steps, whereas Full‑Context often introduces irrelevant information and aggravates context degradation.
In contrast, by allowing the orchestrator to select and compress only task‑relevant history, our method provides a cleaner context and achieves the highest score. 

\subsubsection{Advantage 2: A Learnable Orchestrator}
\label{sec:adv2}

\begin{table}[htbp]
\centering
\small
\setlength{\tabcolsep}{8pt}
\caption{Main results. ReAct is evaluated with a single specified LM per run, while other systems may use either a single LM or a mixed-LM pool. ICL denotes context learning.}
\label{tab:level_main}
\resizebox{\linewidth}{!}{%
\begin{tabular}{l l c c}
\toprule
\textbf{System} & \textbf{LM} & \textbf{Acc.} & \textbf{Avg. Cost} \\
\midrule

\multirow{3}{*}{ReAct}
& Claude-4.5-sonnet  & 53.93 & 0.190 \\
& Claude-4.5-haiku   & 47.88 & 0.066 \\
& Gemini-3-Flash     & 49.09 & 0.070 \\
& GPT-5-mini         & 54.55 & 0.052 \\
& Deepseek-v3.2      & 46.70 & 0.027 \\
\midrule

\multirow{4}{*}{Ours}
& Gemini-3-Flash     & \textbf{80.00} & 0.79 \\
& Claude-4.5-sonnet  & 71.52 & 0.91 \\
& GPT-5-mini         & 67.27 & 0.28 \\
& Deepseek-v3.2      & 67.87 & 0.14 \\

\midrule
Ours (Gemini-3-Flash) & Mixed & 72.12 & 0.70 \\
Ours (ICL)   & Mixed &\textbf{75.15} & \textbf{0.57} \\
\midrule
Ours (Qwen3-8B) & Gemini-3-Flash & 56.97 & 0.36 \\
Ours (SFT) & Gemini-3-Flash & \textbf{68.48}    & \textbf{0.68} \\

\bottomrule
\end{tabular}
}
\end{table}

\paragraph{Supervised fine-tuning (SFT) for task orchestration}
A practical consideration of \our{} is that orchestration quality depends on the main agent’s ability to decompose goals and synthesize high-quality delegation tuples.
To probe this sensitivity, we replace the main agent with a weaker model, \texttt{Qwen3-8B}, while keeping \texttt{Gemini-3-Flash} as the sub-agent executor.
As shown in Table~\ref{tab:level_main}, this setting (\textsc{Ours (Qwen3-8B)}) achieves $56.97\%$ accuracy at \$0.36 average cost. While worse than using a strong main agent (\textsc{Ours} with \texttt{Gemini-3-Flash} reaches $80.00\%$), it still surpasses \texttt{Gemini-3-Flash} with ReAct.
This gap suggests that the orchestration is useful even just with a weak 8B model.
We then fine-tune \texttt{Qwen3-8B} for orchestration via SFT, which yields a large improvement from $56.97\%$ to $68.48\%$ (Table~\ref{tab:level_main}), though this gain comes with an increase of \$0.32 in average cost per task.
By analyzing execution traces, we find that the fine-tuned model exhibits stronger long-horizon problem-solving capabilities when handling complex tasks, increasing the total number of attempts by $56\%$ compared to the base model.
This gain indicates that orchestration is a learnable skill that can be efficiently improved.

\begin{figure}[]
    \centering
    \includegraphics[width=\linewidth]{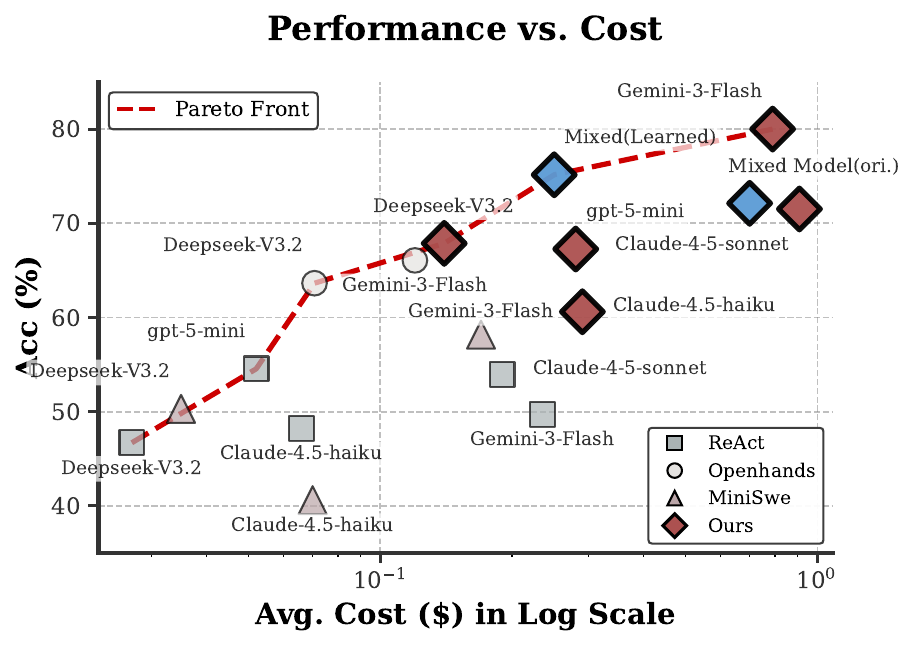}
    \caption{\textbf{Pareto front curve of GAIA.} We plot GAIA accuracy and average cost per task (USD, log scale). Each point corresponds to a configuration, and the dashed curve indicates the Pareto frontier formed by \our across different model routing choices.}
    \label{fig:cost}
\end{figure}

\paragraph{In-context Learning for Cost-aware Orchestration}
Another advantage of \our{} is the ability to balance cost and performance through step-wise model routing.
Table~\ref{tab:level_main} shows that using different sub-agent models leads to markedly different accuracy--cost profiles, making it important for the orchestrator to be sensitive to such trade-offs.
We therefore apply a Pareto-oriented context learning procedure that iteratively optimizes the Orchestrator instruction from interaction trajectories with both performance and monetary cost feedback.
The resulting policy improves \our while reducing cost: under the mixed-model setting, Ours (ICL) improves accuracy from $72.12\%$ to $75.15\%$ while lowering average cost from \$0.70 to \$0.57 (Table~\ref{tab:level_main}), demonstrating that simple prompt-level learning can jointly enhance performance and efficiency.
At the system level, Figure~\ref{fig:cost} further shows that \our naturally yields strong Pareto-efficient operating points: across different model choices, our configurations form the Pareto frontier, indicating a systematically improved cost--performance trade-off over the baselines.

\subsubsection{Advantage 3: Plug-and-play Sub-agents}
\label{sec:adv3}
Here, we aim to verify the framework-level pluggability of our approach. 
Specifically, we replace the execution backend of the sub-agent with different agent frameworks, such as ReAct-style and Mini-SWE-style with \texttt{Gemini-3-Flash} as the orchestrator on Terminal-Bench, following the setups in Appendix~\ref{app:dataset}.

Table~\ref{tab:plugin_subagent} shows that when different sub-agent backends are used, \our maintains stable performance and consistently outperforms the corresponding baselines. 
This design allows sub-agents to be as pluggable modules, enabling the system to remain robust without depending on any particular sub-agent implementation.

\begin{table}[]
\centering
\setlength{\tabcolsep}{6pt}
\caption{Evaluating plug-and-play sub-agents with a fixed orchestrator. We use Gemini-3-Flash to test the robustness and reusability of different sub-agent implementations.
}
\label{tab:plugin_subagent}
\resizebox{\linewidth}{!}{%
\begin{tabular}{l c c c c}
\toprule
\textbf{System} 
& \textbf{Easy} 
& \textbf{Medium}
& \textbf{Hard} 
& \textbf{Acc} \\
\midrule

\multicolumn{5}{l}{\textbf{Standalone baselines}} \\
\midrule
ReAct                     & 50.00 & 34.09 & 16.67 & 28.57 \\
Mini-SWE-Agent            & 50.00 & 40.91& 20.83 & 34.29 \\
Claude Code               & 50.00 & 41.86 & 16.67 & 32.86 \\
\midrule

\multicolumn{5}{l}{\textbf{Orchestrator with plug-in sub-agents}} \\
\midrule
ReAct-style sub-Agent       & 50.00 & \textbf{63.63} & 20.83 & \textbf{48.57} \\
Mini-SWE-style sub-Agent   & \textbf{100.00} & 47.73 & \textbf{33.33} & 44.29 \\
\bottomrule
\end{tabular}
}
\end{table}

\section{Conclusion}
In this work, we present \our{}, an orchestration-centric agentic system that automates sub-agent creation through a unified four-tuple interface (Instruction, Context, Tools, Model) to solve complex, long-horizon agentic tasks. By treating sub-agents as dynamically creatable units, the orchestrator can spawn task-tailored executors on demand with specialized working memory (instruction, context) and capabilities (model, tools). 

This abstraction brings practical benefits: it enables on-demand specialization with task-sufficient context, keeps sub-agents plug-and-play across implementations.
The decoupling of abstraction from execution makes it learnable, and we present two ways to optimize the orchestrator through supervised fine-tuning and context learning. 
Empirically, \our{} demonstrates strong and consistent improvements across three challenging benchmarks (GAIA, Terminal-Bench, and SWE-Bench-Verified) when paired with frontier models like Gemini-3-Flash. It significantly outperforms established baseline frameworks, achieving, for instance, an average gain of 16.28\% points in pass@1 across all benchmarks. These results validate the effectiveness of our orchestration-centric approach to automating complex, long-horizon tasks.

\bibliography{cited}
\bibliographystyle{icml2026}


\clearpage
\onecolumn
\appendix
\label{sec:appendix}

\section{Implemention Details}
\label{app:A}

\subsection{Datasets}
\label{app:dataset}
\begin{itemize}
    \item \textbf{GAIA}~\cite{mialon2023gaia}: 
    GAIA benchmarks general AI assistants on realistic, tool-augmented questions (often involving web browsing and multi-step reasoning).
    We evaluate \our{} and compare with other baselines on the GAIA validation split, which contains a total of 165 tasks.

    \item \textbf{Terminal Bench 2.0}~\cite{tbench_2025}: Terminal-Bench evaluates agents on end-to-end, real-world workflows in a sandboxed command-line environment, graded by executable tests. 
    We evaluate \our{} and compare with other baselines on the Terminal-Bench2.0 test split, which contains a total of 89 tasks. In the main experiments, we randomly sample 70 tasks dut to cost reasons.

    \item \textbf{SWE-Bench-Verified:}~\cite{jimenez2023swe} 
    SWE-Bench Verified measures autonomous software engineering by asking agents to generate patches that resolve real GitHub issues in real repositories, verified by running tests; the Verified split is human-screened to remove problematic cases.
    We evaluate \our{} and compare with other baselines on the SWE-Bench verified version test split, which contains a total of 500 tasks. We randomly sample 100 tasks for evaluation due to cost reasons.
\end{itemize}

\subsection{SFT Hyper-parameters.}
\label{app:sft_hyper}
We use the following hyperparameters during the experiments in Table~\ref{tab:hyper_sft_qwen}.

\begin{table}[htbp]
\centering
\resizebox{0.5\textwidth}{!}{
\begin{tabular}{l|c|l|c|}
\hline
\textbf{Hyperparams} & \textbf{Values} & \textbf{Hyperparams} & \textbf{Values} \\ \hline
learning rate        & 1e-5            & weight decay         & 0.05            \\
warmup ratio         & 0.1             & max length           & 16K            \\
lr scheduler         & cosine          & batch size           & 64              \\
epoch                & 2               & BF16                 & True            \\ 
Deepspeed            & zero3           & tool-call template   & Hermes            \\ \hline
\end{tabular}
}
\caption{SFT Hyperparameters used.}
\label{tab:hyper_sft_qwen}
\end{table}

\subsection{Baseline Implementations}
\label{app:baseline_imp}
For baseline implementations, we evaluate a diverse set of widely-used agentic frameworks. 
During our experiment, we found that \texttt{Claude Code} is not well-suited for GAIA open-world multi-hop question answering due to its architecture and intended usage pattern. Therefore, we did not report the results of \texttt{Claude Code} in the main experiments.

In addition, we find that \texttt{DeepSeek-V3.2} exhibits poor native compatibility with \textsc{Claude Code} based on our initial investigation and empirical trials. Therefore, we exclude this experiment in Table~\ref{tab:main_results}.

For the Terminal-Bench and SWE-Bench evaluations, we leverage the \texttt{Harbor} scaffold to run \texttt{MiniSWE}, \texttt{OpenHands}, and \texttt{Claude Code} under a unified execution interface.

\section{Prompts}
\label{app:prompt_used}

\subsection{Main Agent Prompts}
\label{app:B1}

\subsubsection{GAIA Main Agent Prompt}
\label{app:B1_gaia}

\begin{tcolorbox}[title={\textbf{\small GAIA Main Agent Prompt }}, boxrule=1pt, arc=0mm, colback=black!5!white, colframe=black!75!white, breakable, before skip=10pt, after skip=10pt, pad at break=2mm, parbox=false]

\begin{minted}[fontsize=\scriptsize, breaklines=breakanywhere, frame=lines, framesep=2mm, tabsize=4, style=vs, autogobble]{python}
"""
[GAIA BENCHMARK - QUESTION ANSWERING TASK]
You are the MainAgent (Orchestrator). Your task is to 
solve the given QUESTION by decomposing it into subtasks and delegating each to a sub-agent.

DECISION PROCESS:
1. REVIEW the SUBTASK HISTORY below - check status, result, and key findings of each attempt
2. EVALUATE: Do the results SUFFICIENTLY answer the QUESTION?
   - If any subtask returned a valid result with status "done" → Consider using 'complete'
   - If subtask status is "incomplete" → Review its key findings to see what was accomplished
3. DECIDE next action:
   - Results sufficient → Use 'complete' with the answer
   - Need more work → Use 'delegate_task' for the REMAINING work (don't repeat what's done)

BUDGET AWARENESS:
- You have LIMITED attempts (see Progress below)
- Each delegation costs time and resources - choose models wisely based on task complexity
- If a result looks correct and was verified, trust it and complete

==== MODEL SELECTION GUIDE ====
{model_pricing_table}

Model Selection Strategy:
- Choose cheaper models for simple tasks
- Choose more capable models for complex reasoning or critical attempts

==== Progress ====
[Attempt {attempt_index}/{max_attempts}] Remaining {remaining_attempts} attempts
Budget is limited. Make each attempt count.

==== QUESTION ====
{instruction}

==== SUBTASK HISTORY ====
{subtask_history if subtask_history else "No subtasks completed yet."}

==== AVAILABLE TOOLS ====
{tools_description}


==== OUTPUT ====
ANSWER FORMAT: GAIA requires precise, concise 
answers (single word, number, or short phrase). Do NOT 
include explanations in the answer field.

Return JSON:

If results are SUFFICIENT:
{{
  "action": "complete",
  "reasoning": "The subtask results show [X], which answers the question",
  "params": {{ "answer": "concise answer" }}
}}

If more work is NEEDED:
{{
  "action": "delegate_task", 
  "reasoning": "We have [X] from previous attempts, but still need [Y] to answer the question",
  "params": {{
    "task_instruction": "A SPECIFIC, ACTIONABLE subtask 
    (e.g., 'Extract second word from abstract of paper 2211.xxxxx')",
    "context": "Relevant findings from previous attempts",
    "model": "one of {sub_models}",
    "tools": ["GoogleSearchAction", "ExtractUrlContentAction", 
    "ExecuteCodeAction", "ImageAnalysisAction", "ParseAudioAction"]
  }}
}}
"""
\end{minted}
\end{tcolorbox}

\subsubsection{Terminal-Bench Main Agent Prompt}
\label{app:B1_terminal}

\begin{tcolorbox}[title={\textbf{\small Terminal-Bench Main Agent Prompt }}, boxrule=1pt, arc=0mm, colback=black!5!white, colframe=black!75!white, breakable, before skip=10pt, after skip=10pt, pad at break=2mm, parbox=false]
\begin{minted}[fontsize=\scriptsize, breaklines=breakanywhere, frame=lines, framesep=2mm, tabsize=4, style=vs, autogobble]{python}

"""
You are the MainAgent (Orchestrator). Your task is to 
complete the given software installation/configuration task by delegating to SubAgents.

CRITICAL: CONTAINER LIFECYCLE
- Each SubAgent runs in a FRESH container - if you delegate_task again, the previous work will be lost
- When SubAgent reports status="done", use 'submit' immediately to run tests in that container

==== DECISION PROCESS ====
1. READ the original TASK carefully - identify ALL requirements and edge cases
2. REVIEW SUBTASK HISTORY - check status and completed steps
3. VERIFY SubAgent's work against TASK requirements:
   - Did SubAgent test ALL requirements mentioned in TASK?
   - Did SubAgent test edge cases? (e.g., if TASK mentions "keyboard interrupt", was it actually tested?)
   - Are SubAgent's "completed" items actually addressing the TASK requirements?
4. DECIDE:
   - status="done" AND verification passes → Use 'submit'
   - status="done" BUT verification passes but some requirements are not met → Use 'delegate_task' to fix
   - status="partial" → Use 'delegate_task' with context about what worked/failed

{budget_warning}

==== MODEL SELECTION ====
{model_pricing_table}

==== Progress ====
[Attempt {attempt_index}/{max_attempts}] Remaining {remaining_attempts} attempts

==== TASK ====
{instruction}

==== SUBTASK HISTORY ====
{subtask_history if subtask_history else "No subtasks completed yet."}

==== AVAILABLE TOOLS ====
{tools_description}

==== OUTPUT ====
Return JSON:

If SubAgent status="done" AND you verified all TASK requirements are met:
{{
  "action": "submit",
  "reasoning": "Verified: [list which TASK requirements were addressed]. Submitting.",
  "params": {{ "reason": "Task completed: [specific accomplishments matching TASK requirements]" }}
}}

If SubAgent status="done" BUT verification shows gaps:
{{
  "action": "delegate_task",
  "reasoning": "SubAgent claimed done but [specific gap]: TASK requires [X] but SubAgent only tested [Y]",
  "params": {{
    "task_instruction": "CRITICAL: Previous attempt missed [specific requirement]. 
                        You MUST: [exact steps to fix]",
    "context": " PREVIOUS SUBAGENT CLAIMED DONE BUT MISSED: [specific gap]\\n- 
                 WORKED: [steps to keep]\\n- 
                 MUST FIX: [what was missed]",
    "model": "one of {sub_models}"
  }}
}}

If SubAgent status="partial":
{{
  "action": "delegate_task",
  "reasoning": "SubAgent made partial progress, need to continue with [remaining work]",
  "params": {{
    "task_instruction": "Continue from where previous SubAgent left off: [specific next steps]",
    "context": "From SUBTASK HISTORY:\\n- 
                 WORKED: [steps to REPEAT]\\n- 
                 FAILED: [approaches to AVOID]",
    "model": "one of {sub_models}"
  }}
}}
"""
\end{minted}
\end{tcolorbox}

\subsubsection{SWE-Bench Main Agent Prompt}
\label{app:B1_swe}

\begin{tcolorbox}[title={\textbf{\small SWE-Bench Main Agent Prompt }}, boxrule=1pt, arc=0mm, colback=black!5!white, colframe=black!75!white, breakable, before skip=10pt, after skip=10pt, pad at break=2mm, parbox=false]
\begin{minted}[fontsize=\scriptsize, breaklines=breakanywhere, frame=lines, framesep=2mm, tabsize=4, style=vs, autogobble]{python}

"""You are the MainAgent (Orchestrator) for a SWE-bench task. 
Your goal is to fix a GitHub issue by delegating work to SubAgents.


==== TASK ====
{instruction}

REPOSITORY: {repo}
INSTANCE: {instance_id}

==== DECISION PROCESS ====
1. READ the TASK carefully - understand the GitHub issue and what needs to be fixed
2. REVIEW SUBTASK HISTORY - check SubAgent's progress, completed steps, and test results
3. VERIFY against TASK requirements:
   - Did SubAgent locate the buggy code?
   - Did SubAgent make appropriate code changes?
   - Did SubAgent run tests and confirm the fix works?
4. DECIDE:
   -  status="done" AND tests pass → Use 'submit'
   -  status="done" BUT tests fail or incomplete → Use 'delegate_task' to fix remaining issues
   -  status="partial" → Use 'delegate_task' with guidance on next steps

CRITICAL: SWE-BENCH CONTAINER BEHAVIOR
- When SubAgent reports status="done" with passing tests, use 'submit' to trigger final evaluation
- 'submit' runs the official test suite (FAIL_TO_PASS + PASS_TO_PASS tests) to determine success


==== MODEL SELECTION ====
{model_pricing_table}

==== Progress ====
[Attempt {attempt_index}/{max_attempts}] Remaining {remaining_attempts} attempts
{budget_warning}


==== SUBTASK HISTORY ====
{subtask_history if subtask_history else "No subtasks completed yet."}

==== AVAILABLE TOOLS ====
{tools_description}

==== OUTPUT ====
Return JSON:

If SubAgent status="done" AND tests pass:
{{
  "action": "submit",
  "reasoning": "Verified: [what was fixed, which tests passed]. Submitting for evaluation.",
  "params": {{ "reason": "Fix verified: [specific fix description]" }}
}}

If SubAgent status="done" BUT tests fail or incomplete:
{{
  "action": "delegate_task",
  "reasoning": "SubAgent reported done but [specific issue]: tests show [failure details]",
  "params": {{
    "task_instruction": "CRITICAL: Previous fix incomplete. [specific next steps needed]",
    "context": " ISSUE: [what failed]\\n-  DONE: [completed work]\\n-  TODO: [remaining work]",
    "model": "one of {sub_models}"
  }}
}}

If SubAgent status="partial":
{{
  "action": "delegate_task",
  "reasoning": "SubAgent made partial progress: [summary]. Need to [next steps]",
  "params": {{
    "task_instruction": "Continue: [specific next steps based on SUBTASK HISTORY]",
    "context": "From previous attempt:\\n-  WORKED: [keep these]\\n-  FAILED: [avoid these]",
    "model": "one of {sub_models}"
  }}
}}
"""
\end{minted}
\end{tcolorbox}

\subsection{Sub-Agent Prompts}
\label{app:B2}

\subsubsection{GAIA Sub-Agent Prompt}
\label{app:B2_gaia}

\begin{tcolorbox}[title={\textbf{\small Gaia Sub-Agent Prompt }},   boxrule=1pt, arc=0mm,colback=black!5!white,colframe=darkgray,breakable, before skip=10pt, after skip=10pt,pad at break=2mm, parbox=false]
\begin{minted}[fontsize=\scriptsize, breaklines=breakanywhere, frame=lines, framesep=2mm, tabsize=4, style=vs, autogobble]{python}

ORCHESTRA_GAIA_PROMPT = """
==== Progress ====
[Step {current_step}/{max_steps}] Remaining {remaining_steps} steps
{budget_warning}

==== Your Task (from MainAgent) ====
{task_instruction}

==== Context ====
{context}

==== Original Question (for reference) ====
{original_question}

==== Available Tools ====
{action_space}

==== Guidelines ====
1. Focus on completing YOUR TASK above
2. Think step by step before outputting an action
3. Write key observations to the "memory" field
4. Use print() in ExecuteCodeAction to see computation results
5. Once done, use 'finish' IMMEDIATELY

BUDGET: When remaining_steps <= 5, use 'finish' NOW!

==== Output Format ====
```json
{{
    "action": "<tool_name>",
    "params": {{}},
    "memory": "<observations>"
}}
```

==== Memory ====
{memory}

==== Current Observation ====
{obs}
"""

\end{minted}
\end{tcolorbox}

\subsubsection{Terminal-Bench Sub-Agent Prompt}
\label{app:B2_terminal}

\begin{tcolorbox}[title={\textbf{\small Terminal-Bench Sub-Agent Prompt }},   boxrule=1pt, arc=0mm,colback=black!5!white,colframe=darkgray,breakable, before skip=10pt, after skip=10pt,pad at break=2mm, parbox=false]
\begin{minted}[fontsize=\scriptsize, breaklines=breakanywhere, frame=lines, framesep=2mm, tabsize=4, style=vs, autogobble]{python}

"""
==== Progress ====
[Step {current_step}/{max_steps}] Remaining: {remaining_steps} step(s)
{budget_warning}
If you run out of steps without "finish", your work is lost and marked as timeout.

==== Your Task (from MainAgent) ====
{task_instruction}

==== Context (from previous attempts) ====
{context}
Use this info: repeat what WORKED, avoid what FAILED.

==== Original Question (for reference) ====
{original_question}

==== Action Space ====
{action_space}

==== Memory ====
Recent memory:
{memory}

==== Current Observation ====
{obs}

==== Thinking ====
Think step by step before outputting an action. Write key reasoning in memory for future steps.

==== Action Guidelines ====
You have TWO actions available:

1. **execute** - Run shell commands and observe results
   - Use this to install packages, configure services, verify status, etc.
   - Example: "apt update && apt install -y nginx"

2. **finish** - Report your progress to MainAgent
   - Use when task is COMPLETE (status="done")
   - Use when you made PROGRESS but need more work (status="partial")
   - MUST use before running out of steps! Your work is LOST if you timeout.

**What to report in finish:**
- completed: List SUCCESSFUL steps that WORKED (e.g., ["apt update succeeded", "nginx installed"])
- issues: List FAILED attempts with WHY (e.g., ["nginx -v failed: command not found"])
- message: Brief summary of current state

This info helps the NEXT SubAgent know what to repeat and what to avoid.

==== Output Format ====
CRITICAL: You MUST reply with ONLY a JSON object. No explanations, no markdown, no other text.

For execute:
{{"action": "execute", "params": {{"command": "your shell command"}}, "memory": "key findings"}}

For finish:
{{"action": "finish", "params": {{"status": "done|partial", "completed": [...], 

"issues": [...], "message": "..."}}, "memory": "final notes"}}


"""

\end{minted}
\end{tcolorbox}

\subsubsection{SWE-Bench Sub-Agent Prompt}
\label{app:B2_swe}

\begin{tcolorbox}[title={\textbf{\small SWE-Bench Sub-Agent Prompt }},   boxrule=1pt, arc=0mm,colback=black!5!white,colframe=darkgray,breakable, before skip=10pt, after skip=10pt,pad at break=2mm, parbox=false, before upper=\sloppy]
\begin{minted}[fontsize=\scriptsize, breaklines=breakanywhere, frame=lines, framesep=2mm, tabsize=4, style=vs, autogobble]{python}

SWEBENCH_SUBAGENT_PROMPT = """
You are an autonomous software engineering agent tasked with solving GitHub issues.
You have access to a specialized command interface (ACI) for navigating, viewing, editing, and testing code.
You will work in a Docker container with the repository already cloned and checked out to the correct commit.

==== Progress ====
[Step {current_step}/{max_steps}] Remaining: {remaining_steps} step(s)
{budget_warning}
If you run out of steps without "finish", your work is lost and marked as timeout.

==== Your Task (from MainAgent) ====
{task_instruction}

==== Context (from previous attempts) ====
{context}

==== Current State ====
{state_info}

==== Command Reference ====
{command_docs}

=== FINISH (Report to MainAgent) ===
finish <status> <message>
    Report your progress back to MainAgent. Status MUST be one of:
    - done: Task completed successfully, tests pass
    - partial: Made progress but not finished (e.g., found bug but fix not working)

==== Memory ====
Recent memory:
{memory}

==== Current Observation ====
{observation}

==== OUTPUT FORMAT (STRICT) ====
You MUST output EXACTLY two sections in this order. No other text allowed.

DISCUSSION
<your reasoning here>

COMMAND
<single command here>

RULES:
- DISCUSSION must contain your step-by-step reasoning
- COMMAND must contain exactly ONE command on a single line
- After COMMAND line, do NOT add any explanation, examples, or comments
- Do NOT output anything after the command
"""

\end{minted}
\end{tcolorbox}

\subsubsection{Sub-Agent Summary Prompt}
\label{app:B2_review}

\begin{tcolorbox}[title={\textbf{\small Sub-Agent Summary Prompt }},   boxrule=1pt, arc=0mm,colback=black!5!white,colframe=darkgray,breakable, before skip=10pt, after skip=10pt,pad at break=2mm, parbox=false]
\begin{minted}[fontsize=\scriptsize, breaklines=breakanywhere, frame=lines, framesep=2mm, tabsize=4, style=vs, autogobble]{python}

"""You are a trajectory summarizer. Review the SubAgent's execution trace.
Compare the execution trace against the original task requirements.

== ORIGINAL TASK ==
{original_question}

== EXECUTION TRACE ==
{trace_text}

== OUTPUT ==
Based on the trace, answer:
1. COMPLETED: What requirements from the original task were actually done?
2. REMAINING: What requirements are still missing or not properly tested?

Summarize in 5-10 bullets: key progress, problems, remaining issues.
Output ONLY bullets.Be specific and concise. Output ONLY the two sections above."""

\end{minted}
\end{tcolorbox}

\subsection{Learning Prompt}

\subsubsection{STRATEGY OPTIMIZE PROMPT}
\begin{tcolorbox}[title={\textbf{\small STRATEGY OPTIMIZE PROMPT }},   boxrule=1pt, arc=0mm,colback=black!5!white,colframe=darkgray,breakable, before skip=10pt, after skip=10pt,pad at break=2mm, parbox=false]
\begin{minted}[fontsize=\scriptsize, breaklines=breakanywhere, frame=lines, framesep=2mm, tabsize=4, style=vs, autogobble]{python}

STRATEGY_OPTIMIZE_PROMPT = """
You are optimizing the MainAgent strategy block for GAIA tasks.
Focus on selecting sub-models, deciding when to delegate, and managing cost vs performance.
You need to analysis the model ability and cost on this task, 
and creat a better strategy for main agent to select cheaper model while keep the performance.

Current strategy block:
{strategy}

Evaluation summary:
- pass_rate: {pass_rate}
- avg_reward: {avg_reward}
- total_cost: {total_cost}

Recent trajectories (summarized):
{trajectories}

Write one improved strategy block ONLY. Output in XML:
<prompt>...strategy text...</prompt>
"""

\end{minted}
\end{tcolorbox}

\subsubsection{STRATEGY SELECT PROMPT}

\begin{tcolorbox}[title={\textbf{\small STRATEGY SELECT PROMPT }},   boxrule=1pt, arc=0mm,colback=black!5!white,colframe=darkgray,breakable, before skip=10pt, after skip=10pt,pad at break=2mm, parbox=false]
\begin{minted}[fontsize=\scriptsize, breaklines=breakanywhere, frame=lines, framesep=2mm, tabsize=4, style=vs, autogobble]{python}

STRATEGY_SELECT_PROMPT = """
You are comparing two MainAgent strategy prompts for GAIA tasks.
Summarize each trajectory's strengths/weaknesses, 
relate them to the strategy text,
then decide which strategy is better overall considering BOTH performance and cost.

# A (parent/best-so-far)
pass_rate: {pass_rate_a} / avg_reward: {avg_reward_a} / total_cost: {total_cost_a}
strategy: {strategy_a}
trajectories: {traj_a}

# B (new candidate)
pass_rate: {pass_rate_b} / avg_reward: {avg_reward_b} / total_cost: {total_cost_b}
strategy: {strategy_b}
trajectories: {traj_b}

Respond with XML:
<analysis>your reasoning</analysis>
<choose>A/B</choose>
"""

\end{minted}
\end{tcolorbox}

\section{Case Study}
\label{app:case_study}

\subsection{GAIA Case Study}

\subsubsection{Case Overview}

We evaluate \our{} on 165 tasks from the GAIA benchmark, using Gemini-3-Flash for both the Main Agent and Sub Agent. The results show that \our{} exhibit strong orchestration
capability and robustness in complex, multi-step tasks.
Across all 165 \textsc{GAIA} tasks, \textsc{\our{}} achieves an overall success rate of $80.0\%$. Performance further stratifies by difficulty: Level~1 (easy) reaches $88.7\%$, Level~2 (medium) reaches $80.2\%$, and Level~3 (hard) reaches $61.5\%$. 

\paragraph{Long-Horizon Support and Context Stability.}
The system successfully completes multiple high-cost tasks with long interaction chains. For example, task \texttt{935e2cff} (Level~1) requires 10 attempts with a total cost of \$5.93 yet still completes successfully; task \texttt{853c8244} (Level~2) also succeeds after 10 interaction rounds with a cost of \$3.06. These cases demonstrate that the system can maintain context integrity throughout long-horizon execution, mitigating the ``context rot'' issue commonly observed in standard LLM-based long dialogs.

\paragraph{Strong Error Recovery.}
We find a key strength of \textsc{\our{}} is its self-correction mechanism. On Level~3 hard tasks, \texttt{8131e2c0} and \texttt{0512426f} succeed after 10 attempts, \texttt{983bba7c} succeeds after 9 attempts, and \texttt{872bfbb1} succeeds after 8 attempts. This indicates that even when initial plans fail, the system can progressively converge to the correct answer via Main-Agent reflection and replanning, coupled with iterative execution by Sub Agents. In total, 34 tasks are successfully completed after more than five attempts, highlighting strong error recovery capability.

%

Here we present a representative case study to illustrate how the proposed hierarchical orchestration mechanism operates in practice.

The task is a \textbf{Level-2} question from the GAIA benchmark, asking the agent to identify the 2015 Metropolitan Museum of Art exhibition titled after the Chinese zodiac animal of that year, and count how many figures in the ``twelve animals of the Chinese zodiac'' set have a visible hand. The expected answer is \textbf{11}.

Our orchestrator successfully solved this task in \textbf{10 attempts} through three key phases of iterative refinement:

\textbf{(i) Error Correction via Feedback Loop.} 
In Attempt~1, the main agent initially hypothesized incorrect accession numbers (\texttt{1975.1.784-795}) based on prior knowledge. After the sub-agent reported these were unrelated artworks, the orchestrator \emph{proactively corrected} the hypothesis in Attempt~2, explicitly instructing: \emph{``Do not use 1975.1.784-795 as they are unrelated drawings.''} This demonstrates the system's ability to learn from sub-agent failures and refine task instructions accordingly.

\textbf{(ii) Key Finding Extraction and Propagation.} 
In Attempt~5, although the sub-agent timed out, it discovered a critical piece of evidence: \emph{``The snake's hands are hidden in long, loose sleeves.''} The orchestrator extracted this partial result and persisted it into the context for Attempt~6, writing: \emph{``Previous analysis suggested the snake (02.18.730f) has hands hidden in sleeves.''} This illustrates how our architecture preserves and propagates valuable intermediate findings across attempts.

\textbf{(iii) Hypothesis Formation and Confident Convergence.} 
By Attempt~7, the orchestrator synthesized accumulated evidence to form a concrete hypothesis: the answer is likely 11 (all figures except the snake). Rather than immediately committing, it delegated additional verification tasks to confirm whether any other figures (e.g., monkey, dog, pig) might have ``paws'' instead of ``hands.'' Finally, in Attempt~10, with sufficient corroborating evidence, the orchestrator confidently issued the \texttt{complete} action with the correct answer.

\subsubsection{Detailed Case}

Below is the detailed parameters of Main Agent decision of each attempt:
\begin{tcolorbox}[
  title={\textbf{\small Case Metadata (JSON)}},
  boxrule=1pt, arc=0mm,
  colback=gray!5, colframe=gray!70,
  breakable, before skip=8pt, after skip=10pt,
  pad at break=2mm, parbox=false
]
\begin{minted}[fontsize=\scriptsize, breaklines=true, frame=lines, framesep=2mm, tabsize=4, style=vs, autogobble]{json}
{
  "task_id": "853c8244-429e-46ca-89f2-addf40dfb2bd",
  "question": "In the 2015 Metropolitan Museum of Art exhibition titled after 
  the Chinese zodiac animal of 2015, how many of the \"twelve animals of the Chinese zodiac\"
  have a hand visible?",
  "expected_answer": "11",
  "main_model": "gemini-3-flash-preview",
  "success": true,
  "total_cost": 3.0551280000000003,
  "attempts": 10
}
\end{minted}
\end{tcolorbox}

\label{app:C1_gaia}

%

\begin{tcolorbox}[
  title={\textbf{\small Attempt 1 - Main Decision}},
  boxrule=1pt,
  arc=0mm,
  colback=gray!5,
  colframe=gray!70,
  breakable,
  before skip=10pt,
  after skip=10pt,
  pad at break=2mm,
  parbox=false
]
\begin{minted}[fontsize=\scriptsize, breaklines=breakanywhere, frame=lines, framesep=2mm, tabsize=4, style=vs, autogobble]{json}
{
  "action": "delegate_task",
  "reasoning": "I need to identify the specific 2015 Metropolitan Museum 
  of Art exhibition titled after the Chinese zodiac animal of that year 
  and then count how many of the 'twelve animals of the Chinese zodiac' figures in
  that exhibition have visible hands. 
  This requires online research and visual inspection of museum artifacts.",
  "params": {
    "task_instruction": "
    1. Identify the 2015 Metropolitan Museum of Art exhibition titled after the 2015 Chinese zodiac animal 
    (Note: 2015 was the Year of the Ram/Goat/Sheep).
    \n2. In that exhibition, identify the specific set of 
    'twelve animals of the Chinese zodiac' figures mentioned in the question 
    (likely a set of Tang dynasty ceramic figures).
    \n3. Examine the images of these 12 figures from the Metropolitan Museum's online collection 
    (e.g., accession numbers 1975.1.784 through 1975.1.795).
    \n4. Count how many of these 12 anthropomorphic figures have at least one visible hand 
    (fingers or hand structure visible, not tucked inside sleeves or purely animal paws/hooves).
    \n5. Provide the final count as the answer.",
    "context": "The year 2015 was the Year of the Ram/Goat. 
    The exhibition at the Met was likely titled 'Celebrating the Year of the Ram'.
    There is a famous set of 12 zodiac figures in their collection.",
    "model": "model_1"
  }
}
\end{minted}
\end{tcolorbox}

\begin{tcolorbox}[
  title={\textbf{\small Attempt 2 - Main Decision}},
  boxrule=1pt,
  arc=0mm,
  colback=gray!5,
  colframe=gray!70,
  breakable,
  before skip=10pt,
  after skip=10pt,
  pad at break=2mm,
  parbox=false
]
\begin{minted}[fontsize=\scriptsize, breaklines=breakanywhere, frame=lines, framesep=2mm, tabsize=4, style=vs, autogobble]{json}
{
  "action": "delegate_task",
  "reasoning": "The previous attempt correctly identified the exhibition as 'Celebrating the Year of the Ram' 
  (2015) 
  but failed to correctly identify the specific set of zodiac figures 
  and count the visible hands due to incorrect accession numbers and tool failures. 
  I need to find the specific set of 12 zodiac figures featured in that exhibition 
  and visually inspect them for hands.",
  "params": {
    "task_instruction": "1. Identify the specific set of 'twelve animals of the Chinese zodiac' featured 
    in the 2015 Metropolitan Museum of Art exhibition 'Celebrating the Year of the Ram'.
    \n2. Locate the Met Museum accession numbers for this set 
    (likely Tang dynasty figures, possibly 02.18.730a–l or similar).
    \n3. For each of the twelve figures in that specific set, determine how many have a hand 
    (or fingers/claws with hand-like structure) visible outside of their sleeves.
    \n4. Count the number of such figures with visible hands and provide the integer as the answer.",
    "context": "2015 was the Year of the Ram. The exhibition 'Celebrating the Year of the Ram' ran 
    from January 19 to August 23, 2015. 
    Previous attempts suggest exploring accession numbers 02.18.730a–l. 
    Do not use 1975.1.784–795 as they are unrelated drawings.
    Many Tang ceramic zodiac figures have their hands hidden in sleeves;
    the question specifically asks for visible hands.",
    "model": "model_1"
  }
}
\end{minted}
\end{tcolorbox}

\begin{tcolorbox}[
  title={\textbf{\small Attempt 3 - Main Decision}},
  boxrule=1pt,
  arc=0mm,
  colback=gray!5,
  colframe=gray!70,
  breakable,
  before skip=10pt,
  after skip=10pt,
  pad at break=2mm,
  parbox=false
]
\begin{minted}[fontsize=\scriptsize, breaklines=breakanywhere, frame=lines, framesep=2mm, tabsize=4, style=vs, autogobble]{json}
{
  "action": "delegate_task",
  "reasoning": "Previous attempts identified the 2015 exhibition as 'Celebrating the Year of the Ram' 
  and the likely target set as the 'Twelve animals of the Chinese zodiac' (Accession 02.18.730a–l). 
  However, they failed to count the visible hands due to API errors and mismanagement of findings.
  I am delegating to a sub-agent to specifically find the image for 02.18.730a–l 
  and count the figures with visible hands.",
  "params": {
    "task_instruction": "1. Confirm that the artwork titled 'twelve animals of the Chinese zodiac' 
    (likely Accession 02.18.730a–l) was featured in the 2015 Metropolitan Museum exhibition 
    'Celebrating the Year of the Ram'.
    \n2. For this specific set of 12 figures, identify how many of the animals have a hand visible 
    (i.e., fingers or hand structure visible outside of sleeves, possibly holding an object or resting). 
    \n3. Look for high-resolution images of this set on the Met Museum's website 
    or associated catalogue records to ensure an accurate count for all 12 figures.
    \n4. Provide only the final count as the answer.",
    "context": "The exhibition is 'Celebrating the Year of the Ram' (Jan-Aug 2015). 
    The set is jade figures, Qing dynasty, accession 02.18.730a–l. Previous attempts suggested 
    these are seated anthropomorphic figures. 
    Be careful to check if the hands are truly visible or tucked into the 'hu' tablets or sleeves.",
    "model": "model_1"
  }
}
\end{minted}
\end{tcolorbox}

\begin{tcolorbox}[
  title={\textbf{\small Attempt 4 - Main Decision}},
  boxrule=1pt,
  arc=0mm,
  colback=gray!5,
  colframe=gray!70,
  breakable,
  before skip=10pt,
  after skip=10pt,
  pad at break=2mm,
  parbox=false
]
\begin{minted}[fontsize=\scriptsize, breaklines=breakanywhere, frame=lines, framesep=2mm, tabsize=4, style=vs, autogobble]{json}
{
  "action": "delegate_task",
  "reasoning": "Previous attempts identified the exhibition 'Celebrating the Year of the Ram' 
  and the jade zodiac set (02.18.730a–l) but failed to provide a definitive count of figures with visible hands. 
  I need a sub-agent to specifically analyze the images of the 12 individual figures in the jade set 
  (02.18.730a through 02.18.730l) 
  and count how many have at least one hand visible (fingers or hand structure outside the sleeve).",
  "params": {
    "task_instruction": "1. Confirm the set of 'twelve animals of the Chinese zodiac'
    featured in the Met Museum exhibition 'Celebrating the Year of the Ram' (2015). 
    Identify its accession numbers (likely 02.18.730a–l).
    \n2. For each of the twelve figures in this specific jade set, examine high-resolution images 
    or detailed descriptions on the Metropolitan Museum website 
    to determine if a hand (fingers or hand structure) is visible.
    \n3. Note: A 'visible hand' means the hand is not completely hidden inside a voluminous sleeve.
    It may be holding an object (like a fan, scroll,
    or medicinal herb).
    \n4. Provide the total number (integer) of the twelve figures that have a hand visible.",
    "context": "The exhibition is 'Celebrating the Year of the Ram' (Jan 2015 - Jan 2016). 
    The jade set is Accession No. 02.18.730a–l.
    Previous attempts confirm this set was in the exhibition.
    You need to perform a visual or 
    descriptive audit of each piece (A through L) to count visible hands.",
    "model": "model_1"
  }
}
\end{minted}
\end{tcolorbox}

\begin{tcolorbox}[
  title={\textbf{\small Attempt 5 - Main Decision}},
  boxrule=1pt,
  arc=0mm,
  colback=gray!5,
  colframe=gray!70,
  breakable,
  before skip=10pt,
  after skip=10pt,
  pad at break=2mm,
  parbox=false
]
\begin{minted}[fontsize=\scriptsize, breaklines=breakanywhere, frame=lines, framesep=2mm, tabsize=4, style=vs, autogobble]{json}
{
  "action": "delegate_task",
  "reasoning": "Previous attempts have correctly identified the target set as 
  the Qing dynasty jade 'Twelve animals of the Chinese zodiac' (Accession 02.18.730a–l), 
  which was featured in the 2015 exhibition 'Celebrating the Year of the Ram'. 
  However, they failed to produce a final count of figures with visible hands. 
  I will delegate a task to perform a definitive count by examining the high-resolution images 
  or detailed descriptions of each of the twelve figures (a through l).",
  "params": {
    "task_instruction": "
    1. Access the Metropolitan Museum's collection page for accession number 02.18.730 
    (Object ID 42102) and its sub-items (02.18.730a through 02.18.730l).
    \n2. For each of the twelve figures in this set, determine if a hand 
    (fingers, hand structure, or paws clearly acting as hands) is visible outside of the robe sleeves.
    Often these figures are holding objects like fans, scrolls, or brushes, which makes the hand visible.
    \n3. Count the total number of figures out of the twelve that have at least one hand visible.
    \n4. Provide the final integer count as the answer.",
    "context": "The 2015 Met exhibition is 'Celebrating the Year of the Ram'. 
    The relevant set of 'twelve animals' is 02.18.730a–l. Use high-resolution images like DT5058.jpg 
    (found on the Met website) 
    or individual item photos to verify the visibility of hands for each figure 
    (Rat, Ox, Tiger, Rabbit, Dragon, Snake, Horse, Ram, Monkey, Rooster, Dog, Pig).
    Note that 'visible' means the hand is not completely tucked into a sleeve.",
    "model": "model_1"
  }
}
\end{minted}
\end{tcolorbox}

\begin{tcolorbox}[
  title={\textbf{\small Attempt 6 - Main Decision}},
  boxrule=1pt,
  arc=0mm,
  colback=gray!5,
  colframe=gray!70,
  breakable,
  before skip=10pt,
  after skip=10pt,
  pad at break=2mm,
  parbox=false
]
\begin{minted}[fontsize=\scriptsize, breaklines=true, frame=lines, framesep=2mm, tabsize=4, style=vs, autogobble]{json}
{
  "action": "delegate_task",
  "reasoning": "Previous attempts have correctly identified 
  the 'Twelve animals of the Chinese zodiac' set as the jade figures 
  with accession numbers 02.18.730a–l (Object ID 42102) from the Heber R. 
  Bishop Collection, featured in the 2015 exhibition 'Celebrating the Year of the Ram'. 
  However, they failed to definitively count how many have visible hands.
  Attempt 5 noted that the snake (02.18.730f) has hands hidden in sleeves. 
  I need a sub-agent to find the detailed descriptions for each of the 12 figures (a through l) 
  to determine the visibility of their hands (fingers or hand structure visible, often while holding an object).",
  "params": {
    "task_instruction": "Identify the set of 12 jade Chinese zodiac figures (Accession 02.18.730a–l)
    in the Metropolitan Museum of Art's collection. 
    For each figure 
    (Rat, Ox, Tiger, Rabbit, Dragon, Snake, Horse, Ram, Monkey, Rooster, Dog, Pig),
    determine if at least one hand (or fingers) is visible outside of the sleeves. 
    Many of these anthropomorphic figures hold objects (e.g., scrolls, fans, fly whisks).
    Count the total number of figures where a hand is visible and provide the integer.",
    "context": "The figures are from the Qing dynasty (Qianlong period) 
    and were part of the 2015 exhibition 'Celebrating the Year of the Ram'.
    Accession numbers are 02.18.730a through 02.18.730l. 
    Previous analysis suggested the snake (02.18.730f) has hands hidden in sleeves. 
    Focus on finding individual descriptions or high-res images to check each of the 12.",
    "model": "model_1"
  }
}
\end{minted}
\end{tcolorbox}

\begin{tcolorbox}[
  title={\textbf{\small Attempt 7 - Main Decision}},
  boxrule=1pt,
  arc=0mm,
  colback=gray!5,
  colframe=gray!70,
  breakable,
  before skip=10pt,
  after skip=10pt,
  pad at break=2mm,
  parbox=false
]
\begin{minted}[fontsize=\scriptsize, breaklines=breakanywhere, frame=lines, framesep=2mm, tabsize=4, style=vs, autogobble]{json}
{
  "action": "delegate_task",
  "reasoning": "We have identified the exhibition as 'Celebrating the Year of the Ram' (2015)
  and the centerpiece as the jade set of twelve zodiac animals (Accession 02.18.730a–l).
  A commonly found description states that 'with the exception of the snake,
  whose hands are hidden in its sleeves, each of the other eleven figures holds an object.' 
  This strongly suggests the answer is 11.
  However, the question specifies a 'hand visible' meaning 'fingers or hand structure visible,
  not... purely animal paws/hooves.' 
  I need a final verification that each of the 11 non-snake figures indeed shows human-like fingers 
  or hands rather than animal paws, especially the monkey, dog, and pig.",
  "params": {
    "task_instruction": "1. Verify the specific attributes and hand visibility 
    for the eleven figures in the Met's jade
    'Twelve animals of the Chinese zodiac' (Accession 02.18.730a–l), excluding the snake (02.18.730f).
    \n2. For each of the animals (Rat, Ox, Tiger, Rabbit, Dragon, Horse, Ram, Monkey, Rooster, Dog, and Pig),
    confirm if at least one human-like hand (fingers or hand structure) is visible while holding 
    their respective objects (candle, fan, shield, etc.).
    \n3. Specifically check if any of these anthropomorphic figures are depicted with 'paws' or 'hooves' 
    that would disqualify them under the 'hand visible' definition, 
    or if any have their hands hidden despite holding an object.\n4. 
    Confirm the final count of animals with at least one hand visible.",
    "context": "The exhibition is 'Celebrating the Year of the Ram' (2015). The set is 02.18.730a–l. 
    The snake definitely has no hands visible. 
    Images and descriptions suggest the other 11 hold objects and are in human form.",
    "model": "model_1"
  }
}
\end{minted}
\end{tcolorbox}

\begin{tcolorbox}[
  title={\textbf{\small Attempt 8 - Main Decision}},
  boxrule=1pt,
  arc=0mm,
  colback=gray!5,
  colframe=gray!70,
  breakable,
  before skip=10pt,
  after skip=10pt,
  pad at break=2mm,
  parbox=false
]
\begin{minted}[fontsize=\scriptsize, breaklines=breakanywhere, frame=lines, framesep=2mm, tabsize=4, style=vs, autogobble]{json}
{
  "action": "delegate_task",
  "reasoning": "Previous attempts have correctly identified the target artwork 
  as the set of twelve jade zodiac figures (accession 02.18.730a–l) 
  in the 2015 Metropolitan Museum exhibition 'Celebrating the Year of the Ram'. 
  It has been established that the snake figure (02.18.730f) has its hands hidden in its sleeves. 
  I need to confirm if the other 11 figures all have visible hands while holding their respective objects, 
  or if any others also have hidden hands/paws.",
  "params": {
    "task_instruction": "1. Verify the hand visibility for each of the 12 figures in 
    the Metropolitan Museum's jade set 'Twelve animals of the Chinese zodiac' (Accession 02.18.730a–l).
    \n2. Search for the specific museum description or catalog entry 
    for this set that details which animals have hidden hands. 
    \n3. Specifically, confirm if there is a statement that 'the snake is the only one' with hidden hands, 
    or if other figures (like the pig or dragon) also have obscured hands.
    \n4. Count the number of figures where at least one hand 
    (or finger/human-like hand structure) is visible outside the sleeves.
    \n5. Provide the final integer count.",
    "context": "The exhibition is 'Celebrating the Year of the Ram' (2015). 
    The set is 02.18.730a–l. Previous findings suggest the snake (02.18.730f) has tucked hands.
    We need to be certain about the other 11.",
    "model": "model_1"
  }
}
\end{minted}
\end{tcolorbox}

\begin{tcolorbox}[
  title={\textbf{\small Attempt 9 - Main Decision}},
  boxrule=1pt,
  arc=0mm,
  colback=gray!5,
  colframe=gray!70,
  breakable,
  before skip=10pt,
  after skip=10pt,
  pad at break=2mm,
  parbox=false
]
\begin{minted}[fontsize=\scriptsize, breaklines=breakanywhere, frame=lines, framesep=2mm, tabsize=4, style=vs, autogobble]{json}
{
  "action": "delegate_task",
  "reasoning": "Previous attempts identified the exhibition as 'Celebrating the Year of the Ram' (2015) 
  and the set as the jade 'Twelve animals of the Chinese zodiac' (Accession 02.18.730a–l). 
  Findings suggest the snake (02.18.730f) has its hands hidden in sleeves, while the others hold attributes. 
  I need to confirm if the snake is the ONLY one with hidden hands to arrive at the count (12 - 1 = 11) 
  or if others like the dragon or pig also have hidden hands.",
  "params": {
    "task_instruction": "Identify the set of 12 jade Chinese zodiac figures involved in the 2015 MET exhibition 
    'Celebrating the Year of the Ram' (Accession 02.18.730a–l). 
    For each of the twelve figures, determine if a hand is visible. 
    Specifically, search for the official Met Museum description 
    (possibly in the Bishop Collection catalog) 
    that states which figures have their hands hidden in sleeves 
    and which have visible hands holding attributes. 
    Confirm if the snake is the only one with hidden hands.",
    "context": "The figures are Qing dynasty jade anthropomorphic zodiac animals. 
    Previous attempts found that the snake (02.18.730f) has hidden hands.
    Most sources imply the other 11 hold objects with visible hands.
    Verify if '11' is the correct count by checking if any other animal 
    (e.g., Dragon, Dog, Pig) also has hidden hands.",
    "model": "model_1"
  }
}
\end{minted}
\end{tcolorbox}

\begin{tcolorbox}[
  title={\textbf{\small Attempt 10 - Main Decision (Final)}},
  boxrule=1pt,
  arc=0mm,
  colback=black!5,
  colframe=black!70,
  breakable,
  before skip=10pt,
  after skip=10pt,
  pad at break=2mm,
  parbox=false
]
\begin{minted}[fontsize=\scriptsize, breaklines=breakanywhere, frame=lines, framesep=2mm, tabsize=4, style=vs, autogobble]{json}
{
  "action": "complete",
  "reasoning": "The 2015 Metropolitan Museum of Art exhibition titled after the Chinese zodiac animal of 2015 
  (the Ram) was 'Celebrating the Year of the Ram'. 
  A highlight of this exhibition was the 'Twelve animals of the Chinese zodiac' (Accession 02.18.730a–l),
  a set of 12 Qing-dynasty jade anthropomorphic figures. 
  According to museum descriptions and high-resolution images (such as DT5058.jpg),
  each of the twelve figures holds an attribute (e.g., a candle, fan, scroll, or peach), 
  which requires at least one hand to be visible outside of the sleeves, except for the snake (02.18.730f). 
  The snake is specifically described as sitting with its hands tucked into long, loose sleeves 
  (a scholar's posture), making its hands hidden. 
  Therefore, 11 of the 12 animals have a hand visible.",
  "params": {
    "answer": "11"
  }
}
\end{minted}
\end{tcolorbox}



\section{Tools And Action Space}
\label{app:action_space}


This appendix details the action space of the \our{} framework, including the actions available to the Main Agent and Sub-Agents, as well as the tool inventory and execution constraints for each benchmark.

\subsection{Main Agent Action Space}

The Main Agent is responsible for global task planning and subtask delegation. Its action space consists of two core actions.

\paragraph{\texttt{delegate\_task}.}
This action delegates a well-scoped subtask to a specialized Sub-Agent. The parameter schema is defined as follows:

\begin{tcolorbox}[title={\textbf{\small Action Schema: delegate\_task}},
  boxrule=1pt, arc=0mm,
  colback=gray!5,
  colframe=gray!70,
  breakable, before skip=6pt, after skip=10pt,
  pad at break=2mm, parbox=false]
\begin{minted}[fontsize=\scriptsize, breaklines=breakanywhere, frame=lines, framesep=2mm, tabsize=4, style=vs, autogobble]{json}
{
  "type": "object",
  "properties": {
    "task_instruction": {
      "type": "string",
      "description": "Detailed, actionable instruction for the Sub-Agent"
    },
    "context": {
      "type": "string",
      "description": "Additional context distilled from prior attempts"
    },
    "model": {
      "type": "string",
      "description": "Model alias to use",
      "enum": ["model_1", "model_2", "..."]
    },
    "tools": {
      "type": "array",
      "items": { "type": "string" },
      "description": "Optional subset of tools exposed to the Sub-Agent"
    }
  },
  "required": ["task_instruction", "model"]
}
\end{minted}
\end{tcolorbox}

\paragraph{\texttt{complete}.}
This action submits the final answer and terminates the task.

\begin{tcolorbox}[title={\textbf{\small Action: complete}},
  boxrule=1pt, arc=0mm,
  colback=gray!5,
  colframe=gray!70,
  breakable, before skip=6pt, after skip=10pt,
  pad at break=2mm, parbox=false]
\begin{minted}[fontsize=\scriptsize, breaklines=breakanywhere, frame=lines, autogobble]{json}
{
  "action": "complete",
  "params": {
    "answer": "<final answer string>"
  }
}
\end{minted}
\end{tcolorbox}

\subsection{Sub-Agent Tools For each Benchmark}
\label{app:tools}
Table~\ref{tab:D_tools} summarizes the tools available to Sub-Agents in each benchmark and their corresponding constraints.

\begin{table}[t]
\centering
\small
\caption{Tool inventory and constraints per benchmark.}
\label{tab:D_tools}
\begin{tabular}{l l p{5.8cm}}
\toprule
\textbf{Benchmark} & \textbf{Tool Name} & \textbf{Constraints / Notes} \\
\midrule
\multirow{6}{*}{GAIA}
  & \texttt{GoogleSearchAction} & Serper API; max 5 results; 30s timeout \\
  & \texttt{ExtractUrlContentAction} & Jina API; chunked for long pages; 50s timeout \\
  & \texttt{ExecuteCodeAction} & Sandboxed in \texttt{workspace/temp}; 10s timeout \\
  & \texttt{ImageAnalysisAction} & Vision LLM backend; supports URL \& local files \\
  & \texttt{ParseAudioAction} & Audio-capable LLM backend; multi-format support \\
  & \texttt{finish} & Reports result/status to the Main Agent \\
\midrule
\multirow{3}{*}{Terminal-Bench}
  & \texttt{execute} & Shell commands in Docker/E2B sandbox \\
  & \texttt{finish} & Reports progress without triggering tests \\

\midrule
\multirow{4}{*}{SWE-Bench}
  & \texttt{execute} & Shell commands in a Docker container \\
  & \texttt{view\_file} & Reads a file with line-range specification \\
  & \texttt{edit\_file} & File editing via string replacement \\
  & \texttt{finish} &  Reports progress to main agent \\
\bottomrule
\end{tabular}
\end{table}

\subsubsection{GAIA Tools}

In the GAIA benchmark, Sub-Agents are equipped with tools for web retrieval, code execution, and multimodal analysis.

\begin{itemize}
\item \textbf{\texttt{GoogleSearchAction}.} Performs web search via the Serper API.
\begin{tcolorbox}[colback=gray!5,colframe=gray!70,arc=0mm,boxrule=1pt]
\begin{minted}[fontsize=\scriptsize,breaklines=breakanywhere]{json}
{"action":"GoogleSearchAction",
 "params":{"query":"...","k":5,"gl":"us","hl":"en"}}
\end{minted}
\end{tcolorbox}

\item \textbf{\texttt{ExtractUrlContentAction}.} Extracts webpage content via the Jina API.
\begin{tcolorbox}[colback=gray!5,colframe=gray!70,arc=0mm,boxrule=1pt]
\begin{minted}[fontsize=\scriptsize,breaklines=breakanywhere]{json}
{"action":"ExtractUrlContentAction",
 "params":{"url":"...","browse_query":"..."}}
\end{minted}
\end{tcolorbox}

\item \textbf{\texttt{ExecuteCodeAction}.} Executes Python or Bash code in a sandboxed environment.
\begin{tcolorbox}[colback=gray!5,colframe=gray!70,arc=0mm,boxrule=1pt]
\begin{minted}[fontsize=\scriptsize,breaklines=breakanywhere]{json}
{"action":"ExecuteCodeAction",
 "params":{"code":"...","code_type":"python|bash","timeout_sec":10}}
\end{minted}
\end{tcolorbox}

\item \textbf{\texttt{ImageAnalysisAction}.} Calls a vision-capable LLM backend to analyze images.
\begin{tcolorbox}[colback=gray!5,colframe=gray!70,arc=0mm,boxrule=1pt]
\begin{minted}[fontsize=\scriptsize,breaklines=breakanywhere]{json}
{"action":"ImageAnalysisAction",
 "params":{"query":"...","image_path":"..."}}
\end{minted}
\end{tcolorbox}

\item \textbf{\texttt{ParseAudioAction}.} Calls an audio-capable LLM backend to process audio inputs.
\begin{tcolorbox}[colback=gray!5,colframe=gray!70,arc=0mm,boxrule=1pt]
\begin{minted}[fontsize=\scriptsize,breaklines=breakanywhere]{json}
{"action":"ParseAudioAction",
 "params":{"query":"...","audio_path":"..."}}
\end{minted}
\end{tcolorbox}

\item \textbf{\texttt{finish}.} Reports subtask results back to the Main Agent.
\begin{tcolorbox}[colback=gray!5,colframe=gray!70,arc=0mm,boxrule=1pt]
\begin{minted}[fontsize=\scriptsize,breaklines=breakanywhere]{json}
{"action":"finish",
 "params":{"result":"...","status":"done|partial|blocked","summary":"..."}}
\end{minted}
\end{tcolorbox}
\end{itemize}

\subsubsection{Terminal-Bench Tools}

In Terminal-Bench, Sub-Agents execute shell commands inside Docker/E2B sandboxes using the following tools:
\begin{itemize}
    \item \textbf{\texttt{execute}:} run shell commands and return outputs.
    \item \textbf{\texttt{finish}:} report intermediate progress without triggering tests.
\end{itemize}

\subsubsection{SWE-Bench Tools}

In SWE-Bench, Sub-Agents are equipped with code navigation and editing capabilities:
\begin{itemize}
    \item \textbf{\texttt{execute}:} run shell commands (e.g., \texttt{git} operations and tests).
    \item \textbf{\texttt{view\_file}:} read file content with a specified line range.
    \item \textbf{\texttt{edit\_file}:} edit files via string replacement.
    \item \textbf{\texttt{finish}:} Report your progress back to MainAgent.
\end{itemize}

\subsection{Sandbox and Network Constraints}

\paragraph{Code Execution Sandbox.}
\texttt{ExecuteCodeAction} executes code in an isolated directory (\texttt{workspace/temp}). Potentially destructive Bash operations are disallowed, including file deletion, privilege escalation, permission changes, root-level redirection, and system-level commands. The default execution timeout is 10 seconds.

\paragraph{Network Constraints.}
Web access is mediated exclusively through tool APIs. Web search is performed via the Serper API (\texttt{google.serper.dev}) with a 30-second timeout, while URL content extraction is handled via the Jina API (\texttt{r.jina.ai}) with a 50-second timeout. Raw HTTP requests are not directly exposed to agents.

\paragraph{Terminal-Bench Sandbox.}
Terminal-Bench supports Docker, E2B, and Daytona backends. The default execution timeout is 600 seconds, and the working directory is automatically inferred from the Dockerfile \texttt{WORKDIR} directive.

\paragraph{SWE-Bench Sandbox.}
Each SWE-Bench task runs in an isolated Docker container. The system automatically clones the target repository and checks out the specified base commit. Tests are executed with \texttt{pytest} under a 300-second timeout.

\section{Third-Party API and Model Pricing}
\label{app:E}

\subsection{Third-Party APIs}
\label{app:E1}

We rely on a small set of third-party APIs to support web search and sandboxed agent environment creation/execution.

\begin{table}[h]
\centering
\small
\begin{threeparttable}
\caption{Third-party APIs used in our system.}
\label{tab:third_party_apis}
\begin{tabular}{l l p{0.56\linewidth}}
\toprule
\textbf{API} & \textbf{Role} & \textbf{How it is used in \our{}} \\
\midrule
\textbf{Serper} (\texttt{serper.dev}) & Web search & Used as the primary search API for retrieving relevant webpages/snippets during GAIA-style information-seeking subtasks. \\
\textbf{Jina} (\texttt{jina.ai}) & Web content retrieval & Used for lightweight webpage fetching/reading (e.g., converting a URL into clean text for extraction) to support search-and-read subtasks. \\
\textbf{E2B} (\texttt{e2b.dev}) & Sandbox environment & Used to create isolated execution environments for agent tool use (e.g., running code or environment-dependent operations) with controlled resources. \\
\bottomrule
\end{tabular}
\begin{tablenotes}[flushleft]\footnotesize
\item \textbf{Usage.} These APIs are invoked only through our tool interface; the main agent and sub-agents do not access external services directly.
\item \textbf{Reproducibility.} When applicable, we cache retrieved web content and log API responses/metadata (e.g., timestamps, query strings, and URLs) to ensure consistent evaluation.
\end{tablenotes}
\end{threeparttable}
\end{table}

\subsection{Model Pricing Table and Cost Accounting}
\label{app:E2}


\begin{tcolorbox}[title={\textbf{\small Cost Table }}, boxrule=1pt, arc=0mm, colback=black!5!white, colframe=black!75!white, breakable, before skip=10pt, after skip=10pt, pad at break=2mm, parbox=false]
\begin{minted}[fontsize=\scriptsize, breaklines=breakanywhere, frame=lines, framesep=2mm, tabsize=4, style=vs, autogobble]{text}

    PRICES = {
        # openai: https://platform.openai.com/docs/pricing
        # anthropic: https://docs.anthropic.com/en/docs/about-claude/pricing
        # openrouter: https://openrouter.ai/
        "gpt-4o-mini": {"input": 0.00015, "output": 0.0006},
        "o3": {"input": 0.002, "output": 0.008},
        "o3-mini": {"input": 0.0011, "output": 0.0044},
        "gpt-5": {"input": 0.00125, "output": 0.01},
        "gpt-5-mini": {"input":0.00025, "output": 0.002},
        "claude-sonnet-4-20250514": {"input": 0.003, "output": 0.015},
        "moonshotai/kimi-k2": {"input": 0.000296, "output": 0.001185}, 
        "deepseek/deepseek-chat-v3.1": {"input":0.00025 , "output":0.001},
        "deepseek-chat": {"input":0.00025 , "output":0.001},
        "z-ai/glm-4.5": {"input": 0.00033, "output": 0.00132},
        "gemini-2.5-pro": {"input": 0.00125, "output": 0.01},
        "claude-4-sonnet": {"input": 0.003, "output": 0.015},
        "claude-sonnet-4-5": {"input": 0.003, "output": 0.015},
        "claude-4-5-haiku": {"input": 0.00088, "output": 0.0044},
        "claude-4-sonnet-20250514": {"input": 0.003, "output": 0.015},
        "gemini-2.5-flash": {"input": 0.0003, "output": 0.00252},
        "gemini-3-flash-preview": {"input": 0.0005, "output": 0.003},
        "gemini-3-pro-preview": {"input": 0.002, "output": 0.004},
        "gemini-2.5-flash-image": {"input": 0.0003, "output": 0.03},
    }


\end{minted}
/\end{tcolorbox}

\end{document}